\documentclass[
]{ceurart}

\usepackage{listings}
\usepackage{array}
\newcolumntype{P}[1]{>{\centering\arraybackslash}p{#1}}
\newcolumntype{M}[1]{>{\centering\arraybackslash}m{#1}}

\usepackage{tabularx}

\usepackage{amsmath}
\usepackage{placeins}

\usepackage{multirow}
\usepackage{adjustbox}
\usepackage{booktabs}

\usepackage{algorithm}
\usepackage[noend]{algpseudocode}

\usepackage{hyperref}

\usepackage{amssymb}
\usepackage{amsthm}
\newtheorem{definition}{Definition}

\begin{document}

%%
%% Rights management information.
%% CC-BY is default license.
\copyrightyear{2024}
\copyrightclause{Copyright for this paper by its authors. Use permitted under Creative Commons License Attribution 4.0 International (CC BY 4.0)}

%%
%% This command is for the conference information
\conference{EXPLIMED - First Workshop on Explainable Artificial Intelligence for the medical domain - 19-20 October 2024, Santiago de Compostela, Spain}

%%
%% The "title" command
\title{Explaining Bayesian Networks in Natural Language using Factor Arguments. Evaluation in the medical domain.}

% \tnotemark[1]
% \tnotetext[1]{You can use this document as the template for preparing your
%   publication. We recommend using the latest version of the ceurart style.}

%%
%% The "author" command and its associated commands are used to define
%% the authors and their affiliations.
\author[1]{Jaime Sevilla}[%
% orcid=0000-0002-0877-7063,
% email=kulyabov-ds@rudn.ru,
% url=https://yamadharma.github.io/,
]
% \cormark[1]
% \fnmark[1]
\address[1]{University of Aberdeen, Aberdeen, UK}

\author[2]{Nikolay Babakov}[%
% orcid=0000-0002-0877-7063,
email=nikolay.babakov@usc.es,
% url=https://yamadharma.github.io/,
]
\cormark[1]
% \fnmark[1]
\address[2]{Centro Singular de Investigación en Tecnoloxías Intelixentes (CiTIUS), Universidade de Santiago de Compostela, Santiago de Compostela, Galicia, Spain}

\author[1]{Ehud Reiter}[]

\author[2]{Alberto Bugarín}[]

% \fnmark[1]

%% Footnotes
\cortext[1]{Corresponding author.}
% \fntext[1]{These authors contributed equally.}

%%
%% The abstract is a short summary of the work to be presented in the
%% article.
\begin{abstract}
  In this paper, we propose a model for building natural language explanations for Bayesian Network Reasoning in terms of factor arguments, which are argumentation graphs of flowing evidence, relating the observed evidence to a target variable we want to learn about. We introduce the notion of factor argument independence to address the outstanding question of defining when arguments should be presented jointly or separately and present an algorithm that, starting from the evidence nodes and a target node, produces a list of all independent factor arguments ordered by their strength. Finally, we implemented a scheme to build natural language explanations of Bayesian Reasoning using this approach. Our proposal has been validated in the medical domain through a human-driven evaluation study where we compare the Bayesian Network Reasoning explanations obtained using factor arguments with an alternative explanation method. Evaluation results indicate that our proposed explanation approach is deemed by users as significantly more useful for understanding Bayesian Network Reasoning than another existing explanation method it is compared to. 
\end{abstract}

%%
%% Keywords. The author(s) should pick words that accurately describe
%% the work being presented. Separate the keywords with commas.
\begin{keywords}
  Bayesian Networks explanation \sep 
  explainable Artificial Intelligence \sep
  Natural language explanations \sep 
  human evaluation of explanations \sep 
  evaluation in the medical domain
\end{keywords}

%%
%% This command processes the author and affiliation and title
%% information and builds the first part of the formatted document.
\maketitle

%% main text
\section{Introduction}
\label{sec:intro}

   It is generally accepted that a proper explanation of AI models is one of the requirements for trustworthiness~\cite{doran2017does,hagras2018toward,floridi2019establishing}. Whereas the accuracy of an AI model is important in many fields, the inability to explain the reasoning or rationale behind the model may block any perspective on its real-life usage, especially in critical domains. Within AI, Bayesian Networks (BNs) can represent knowledge and perform reasoning in contexts of uncertainty. However, interpreting BNs reasoning may be quite a complex task for users because the reasoning mechanism can run in different directions (e.g., from causes to consequences and vice versa). Moreover, the linkage between variables can lead to complex and indirect relationships, complicating the interpretation.

    % 1) @AlWith the significant progress of Artificial Intelligence (AI) representatives of various fields became interested in using it for real-world tasks. 2) Whereas in some fields specialists may trust the model just based on the good metrics with the data unseen during model training, there are many fields, where the additional reason to trust the model's predictions is necessary to make a decision. 3)This naturally yields the need for a trustworthy AI~\cite{floridi2019establishing}. 4)Proper explanation of the AI model is one of the main prerequisites for trustworthiness~\cite{hagras2018toward,doran2017does}. 5)Bayesian Networks are powerful AI instruments, capable of making predictions in an incomplete information setup and that is also explainable. 
    
\begin{figure*}[t!]
    \centering
    \includegraphics[width=\linewidth]{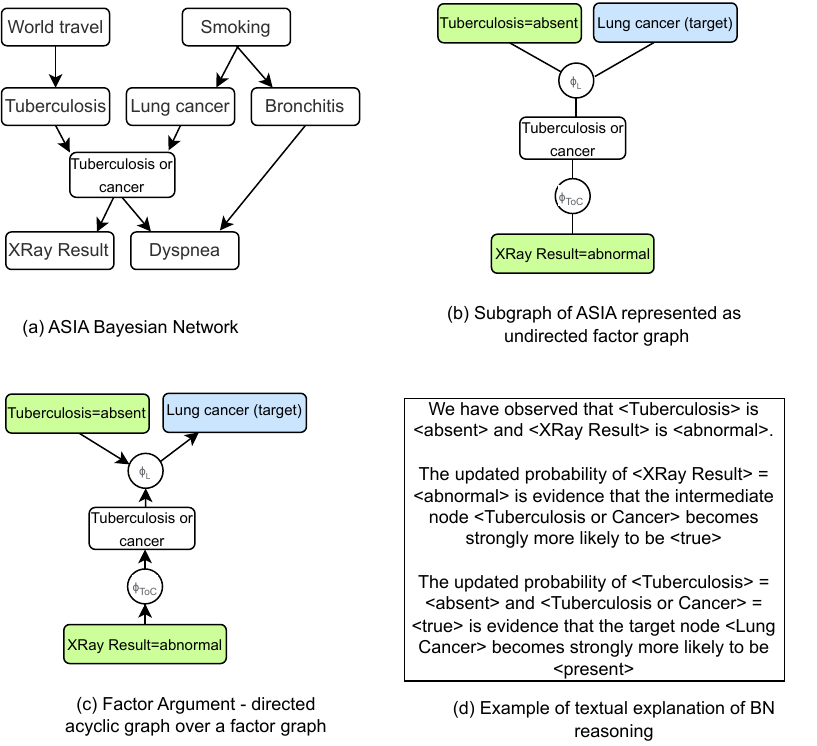}
    \caption{(a) An example of a Bayesian Network is the ASIA Network~\cite{10.2307/2345762}. (b) Visualization of an undirected subgraph of the ASIA BN factor graph, determining the nodes to be used for the reasoning explanation given two evidence nodes (XRay Result = abnormal and Tuberculosis = absent) and a target node Lung cancer. (c) Visualization of Factor Argument: directed acyclic subgraph of a factor graph BN. (d) Possible textual explanation of reasoning.}
    \label{fig:reasoning_sample}
    \end{figure*}

    BNs are directed acyclic graphs whose edges show causal or influential relationships between nodes and that can encode any discrete or continuous variable (see Figure~\ref{fig:reasoning_sample}a for an illustrative example of a BN structure). BNs have been used in various fields: medicine~\cite{mascaro2022modeling}, law~\cite{vlek2016method}, harvesting~\cite{sottocornola2023dssapple}, etc. To complement these efforts, many researchers~\cite{kyrimi2020incremental,keppens2019explainable} are now focusing on studying better ways to present the reasoning of BN to domain experts.

    % The usage of Artificial Intelligence in various fields, such as medicine, has certain limitations, and one of the most significant one is the necessity of explanation. It is natural, that the specialists whose decisions may have significant consequences need to trust the guidance provided by the model~\cite{jamieson2019clinical}, and  Bayesian Network (BN) may be an option for generating such explainable guidance. 

    In general, BN reasoning is presented either visually~\cite{keppens2016explaining} or in natural language~\cite{keppens2019explainable,kyrimi2020incremental}. In this paper, we focus on textual explanations. Arguably, the most important challenge of textual approaches is content determination~\cite{reiter_dale_2000} - the decision on what information (e.g., what variables in the case of BNs) should be used for an explanation. 

    To the best of our knowledge, the existing textual explanation approaches do not explicitly consider the path through which the message passes from the evidence node(s) to the target node. Delivering the explanation using such paths could allow BN users to understand the exact chains of reasoning within a BN that yield the final probability update in the target node. In this paper, we consider a novel content determination strategy that finds the most significant of the aforementioned paths and then textually presents them. We refer to these paths as factor arguments. They are delivered to a BN user in natural language. We compare the proposed approach to existing approaches with human-driven evaluation and open-source our code\footnote{\url{https://gitlab.nl4xai.eu/nikolay.babakov/bn_explanation_with_factor_arguments/}}.

    The rest of the paper is organized as follows. Section~\ref{sec:method} outlines all the details of the proposed method. Section~\ref{sec:proof-of-concept} describes the design and the result of computational experiments aimed to test the computational limitations of the proposed explanation approach.  Section~\ref{sec:human_eval} reports the results of the human evaluation of the method in the medical domain.

\section{Related work}
\label{sec:related}

There are many explanation methods for Bayesian networks~\cite{lacave2002review,hennessy-etal-2020-explaining} that can be delivered in different modalities. 
Druzdzel and Henrion~\cite{henrion1990qualitative} proposed two types of Bayesian network explanations: Qualitative Belief Propagation and Scenario-based Reasoning. Qualitative Belief Propagation focuses on tracing the qualitative effects of evidence through a belief network, emphasizing the direction and impact of evidence from one variable to the next. Scenario-based Reasoning, on the other hand, generates alternative causal stories to account for the evidence, offering a narrative-based approach to understanding the outcomes of probabilistic reasoning. Both approaches aim to enhance the comprehensibility of Bayesian inference, catering to different aspects of human reasoning under uncertainty. Zukerman et al.~\cite{zukerman1996perambulations,zukerman1998bayesian} explain BNs using an iteratively generated argument graph, which consists of a subgraph of said BN. The Elvira tool~\cite{lacave2007explanation} highlights the links within the BN that offer qualitative insight into the conditional probability tables.~\cite{pereira2019content} generates quantified statements and reasons with text using fuzzy syllogism. ~\cite{DEWAAL2022118348} deliver an explanation in a table view, exploiting the generalized Bayes factor score to determine important nodes. ~\cite{koopman2021persuasive} generates contrastive explanations using the annotated lattice obtained by the relations of BN nodes, and ~\cite{renooij2022relevance} uses the concept of Maximum A Posteriori independence to define the nodes relevant to the reasoning explanation.

Our approach delivers the explanation as a sequence of text statements describing the process of BN reasoning. It builds on previous work on extracting chains of reasoning from BNs like the one used in the INSITE system ~\cite{suermondt1992explanation}. It was later refined in~\cite{haddawy1997banter} and~\cite{kyrimi2020incremental}. These approaches suggest how to measure and explain the effect of the available evidence on a target node, but they are quite limited when it comes to explaining interactions between chains of reasoning. We also expand the proposal of \cite{vreeswijk2005argumentation} which explains BNs using argumentation theory. In this work, each conditional probability table is translated into a rule, producing an argumentation scheme that operates similar to Maximum-a-posterior probability inference. We introduce our own formal notion of an argument, that attempts to isolate the effects of different inference paths in the network.
    
In \cite{keppens2012argument} an initial argumentation diagram is refined and labeled to capture the interactions in a BN. This work led us to propose an alternative characterization of argument independence. In~\cite{yap2008explaining} the BN explanatory approach restructures the BN such that the target node has its Markov nodes as parents, then the target CPT is condensed into decision trees, and finally the explanatory nodes under specific contexts are identified by traversing the DTs, and the corresponding BN explanations are generated dynamically. A similar approach was proposed in~\cite{keppens2016explaining,TIMMER2017475} where an argument diagram named support graph (a directed subgraph of the original BN) is extracted from the original BN and further used for explanation. It relates all available evidence to the target in a series of inference steps. The named approaches do not disentangle the separate effects of the different premises of each step on the rule conclusion, as far as the final explanation is delivered as one argument without explicit designation of what particular evidence node or path of belief update made a more significant impact on the final update of the probability. In our work, we address this limitation. 

% Their approach doesn't disentangle the separate effects of the different premises of each step on the rule conclusion; we address this problem.

Unlike modern model-agnostic explainability frameworks like LIME~\cite{ribeiro-etal-2016-trust} and SHAP~\cite{lundberg2017unified}, we focus on explaining which relations between variables are more important rather than explaining which variables are more important overall.

% This difference — that we focus on explaining which relations between variables are more important rather than explaining which variables are more important overall — is also what distinguishes our approach to modern model-agnostic explainability frameworks like LIME~\cite{ribeiro-etal-2016-trust} and SHAP~\cite{lundberg2017unified}.  For example, while SHAP can be useful to detect that bronchitis is important to determine if the patient has dyspnea in Figure \ref{fig:asia_network}, it does not offer a direct answer to the question of whether the path $\text{Bronc} \rightarrow \text{Dysp}$ is more important than the path $\text{Bronc} \leftarrow \text{Smoking} \rightarrow \text{Lung} \rightarrow \text{Either} \rightarrow \text{Dysp}$ to explain the outcome.  

To generate our explanations, we follow some of the guidelines established in~\cite{miller2019explanation}. In particular, we emphasize the need for a selective explanation, offer a contrastive explanation, and show how to explain quantitative arguments in qualitative terms.

\section{A proposal for explaining Bayesian Networks}
\label{sec:method}

% As an illustrative example, consider the BN shown in Figure~\ref{fig:reasoning_sample}a. 
Users normally interact with the BN by entering evidence and querying how the probability of a target variable changes in response. For example, for the BN in Figure~\ref{fig:reasoning_sample}a the user may indicate that the patient does not have Tuberculosis and has abnormal XRay results and may be interested in learning how likely it is that the patient has Lung cancer. 

Our goal in this paper is to propose a model for building natural-language explanations about the reasoning process occurring in a BN. The content to be included in the explanation is determined as a set of directed subgraphs of the BN's factor graph, which we refer to as factor arguments (FAs). Figure~\ref{fig:reasoning_sample}c shows an example of a \textit{FA} relating the evidence nodes `XRay Results' and `Tuberculosis' to the target node `Lung cancer', Figure~\ref{fig:reasoning_sample}d shows a textual description of this \textit{FA}.

In this article, we deal exclusively with BNs with discrete categorical value nodes. Continuous ones fall out of the scope of our work, and while this method can in principle be applied to ordinal values, we do not think our approach is particularly well suited to them.

\subsection{Preliminaries}
\label{sec:prelim}

A BN is a probabilistic graphical model that represents a set of random variables and their conditional dependencies via a directed acyclic graph~\cite{koller2009probabilistic}. Its nodes are associated with conditional probability tables (CPTs), which express the probability of a variable taking a particular value given the outcomes of its direct predecessors.

\begin{table}

\begin{minipage}{0.35\linewidth}
  \centering
  %\caption{Table 1}
  \begin{adjustbox}{width=\linewidth}
    \begin{tabular}{@{}lcc@{}}
      & \multicolumn{2}{c}{Tuberculosis} \\
      \cmidrule(lr){2-3} 
      World Travel & Yes & No \\ \midrule
      Yes & 0.05 & 0.95 \\
      No & 0.01 & 0.99 \\ \bottomrule
    \end{tabular}
  \end{adjustbox}
\end{minipage}
% \hfill
\hspace{0.15\linewidth}
\begin{minipage}{0.35\linewidth}
  \centering
  %\caption{Table 2}
  \begin{adjustbox}{width=\linewidth}
    \begin{tabular}{@{}llc@{}}
      World Travel & Tuberculosis & $\phi_T$ \\ \midrule
      Yes & Yes & 0.05 \\
      Yes & No & 0.95 \\
      No & Yes & 0.01 \\
      No & No & 0.99 \\ \bottomrule
    \end{tabular}
  \end{adjustbox}
\end{minipage}
\caption{An example of CPT of Tuberculosis node from the ASIA Bayesian Network (left) represented as a factor (right).}
\label{tab:factor}
\end{table}

CPTs may be represented as a factor~\cite{koller2009probabilistic}, which is a function mapping all possible values of one or more random variables (its scope) to positive real numbers corresponding to the values in the CPT (see Table~\ref{tab:factor} for an example).  The Scope of a node refers to the set of variables that are directly influenced by or directly influence a given node within a network. Factors can be multiplied, divided, marginalized, and normalized. They can be used to represent a BN as a factor graph - an undirected bipartite graph containing variable and factor nodes. The factor graph only contains edges between variable nodes and factor nodes, and it is parameterized by a set of factors (see Figure~\ref{fig:reasoning_sample}b). Each factor node is associated with precisely one factor, whose scope is the set of variables adjacent to the factor node in the graph. \footnote{For the sake of simplicity, we will use capital letter variables such as \(X,Y,Z\) to refer to variable nodes in a factor graph, and we will use notation such as \(\phi_X\) to refer to the unique factor node that represents the CPT for node \(X\). We will abuse the notation and also refer to \(\phi_X\) as the factor that represents such a CPT.}

Our explanation framework is based on the loopy message passing algorithm for inference in a Bayesian Network~\cite{koller2009probabilistic}. The goal of this algorithm is to approximately compute queries of the form $$P(T=t| E_1=e_1, ..., E_m = e_m)$$ That is, given some evidence $E_i=e_i, i=1, ..., m$ we want to compute how beliefs about a target variable $T$ in the network change. Message passing starts with defining the initial potential factors of each node in the factor graph. The factors of factor nodes are lifted from the respective CPT, and the factors of variable nodes are defined as lopsided factors (where one state probability equals 1 and others 0) if their value is known and as constant factors otherwise. 

The information between the variable nodes is communicated with messages of the form $\mu_{A,B}, \mu_{B,A}$ for each pair of adjacent nodes in the associated factor graph. Each message is initialized as a constant factor and aimed to connect variables of the factor graph in both directions. The factors are updated with the following formula: $$\mu_{A,B} = \sum_{\text{Scope}(A) \setminus \text{Scope}(B)} \phi_A \prod_{C\not = B} \mu_{C,A}$$ Namely, we take the factor $\phi_A$ associated with the sender node $A$, multiply it by all messages sent to $A$, except the one coming from the receiver $B$, and finally marginalize all nodes not in the intersection of scopes of the sender and receiver nodes. \footnote{Note that loopy message passing is a method of approximate inference, which is not guaranteed to converge.}

\subsection{Factor arguments}

While inference algorithms such as message passing will provide an approximately correct conclusion, it is often hard to understand, even for experts, how the evidence influenced the results. To make the BN reasoning process more understandable, we want to deliver to the user a list of considerations explaining how the given pieces of evidence affected the target variable. To do so, we need a way to abstractly represent the considerations that will be turned into explanations. For this purpose, we introduce factor arguments (\textit{FAs}) - webs of flowing evidence relating the evidence nodes and the target node. We represent \textit{FA}s as directed acyclic graphs over a factor graph, which trace the path followed by messages in loopy message propagation from the evidence to a target node.

\begin{definition}[Factor Argument (\textit{FA})]
    A factor argument in a BN is a directed acyclic graph whose skeleton corresponds to a subgraph of the factor graph of said BN. As such, it is composed of alternating variable and factor nodes.

    A factor argument has a single sink, corresponding to a variable node, we will call the target node. Each of its sources is also a variable node, which we will call the evidence node.

\end{definition}

An example of a factor argument is shown in Figure~\ref{fig:reasoning_sample}c.

\subsection{Step effect and factor argument effect}
\label{sec:arg_effect}
% http://localhost:8888/notebooks/bn_nlg/reproduce_algo/expl_bn_small.ipynb for detailed analysis of algo

To define the effect that a \textit{FA} has on the target variable, we introduce the notion of \textit{factor argument effects}. We can understand a \textit{FA} as a series of inference steps, where for each factor node in the \textit{FA} we combine some belief updates about its direct predecessors with the factor itself to produce a belief update for the successor of the factor node. The direct predecessors in \textit{FA} are defined as follows:
% We will now make this process quantitative.

% Definition predfa
\begin{definition}[Direct Predecessors in Factor Argument (\(Pred_{FA}\))]
For any node (either factor node or variable node) within a Factor Argument (FA), the set of direct predecessors, denoted as \(Pred_{FA}(node)\), includes all nodes that have a direct edge leading to the considered node within the FA. Formally, for a variable node \(X\) within an FA:

\begin{equation}
Pred_{FA}(X) = \{\phi \mid \phi \text{ has a direct edge to } X \text{ within the FA}\}
\end{equation}

and for a factor node \(\phi\) :

\begin{equation}
    Pred_{FA}(\phi) = \{X \mid X \text{ has a direct edge to } \phi \text{ within the FA}\}
\end{equation}

\end{definition}

Note that the direct predecessors of a factor node will all be variable nodes, and viceversa. Our goal will be to summarize the flow of evidence between variable nodes as mediated by factor nodes. In other words, we will break down a \textit{FA} in a series of steps, one for each factor in the factor argument, where a belief update on its direct predecessors, i.e., its premises, will produce a belief update on its only successor.

In our formalism, similarly to message passing, beliefs about variables are represented as factors. We use the notation $\delta_X$ to represent a belief update on node $X$, i.e., a factor whose scope is the same as all possible values of $X$. Suppose we are interested in evaluating the effect of factor $\phi$, whose direct predecessors are $\text{Pred}_{FA}(\phi)$, on its successor node $X$. We measure this through the step effect of $\phi$ on $X$.

\begin{definition}[Step Effect (\textit{SE})]
Let $X$ be a variable node in a factor argument $FA$, $\phi$ a factor node that precedes $X$ and $\Delta_{\phi} = \{\delta_Y : Y \in \text{Pred}_{FA}(\phi)\}$ a set of belief updates for each of the variable nodes that precede $\phi$ on $FA$. The step effect (SE) of factor $\phi$ on $X$ given premises $\Delta_\phi$ is defined as:
\begin{equation}SE(\phi, \Delta_{\phi}, X) = 
\frac{\sum\limits_{\text{Pred}_{FA}(\phi)} \phi \cdot \prod\limits_{\delta \in \Delta_{\phi}}\delta }{\sum\limits_{\text{Pred}_{FA}(\phi)} \phi}
\label{eq:step_effect}
\end{equation}
\end{definition}

\begin{figure}[t!]
    \centering
     \includegraphics[width=\linewidth]{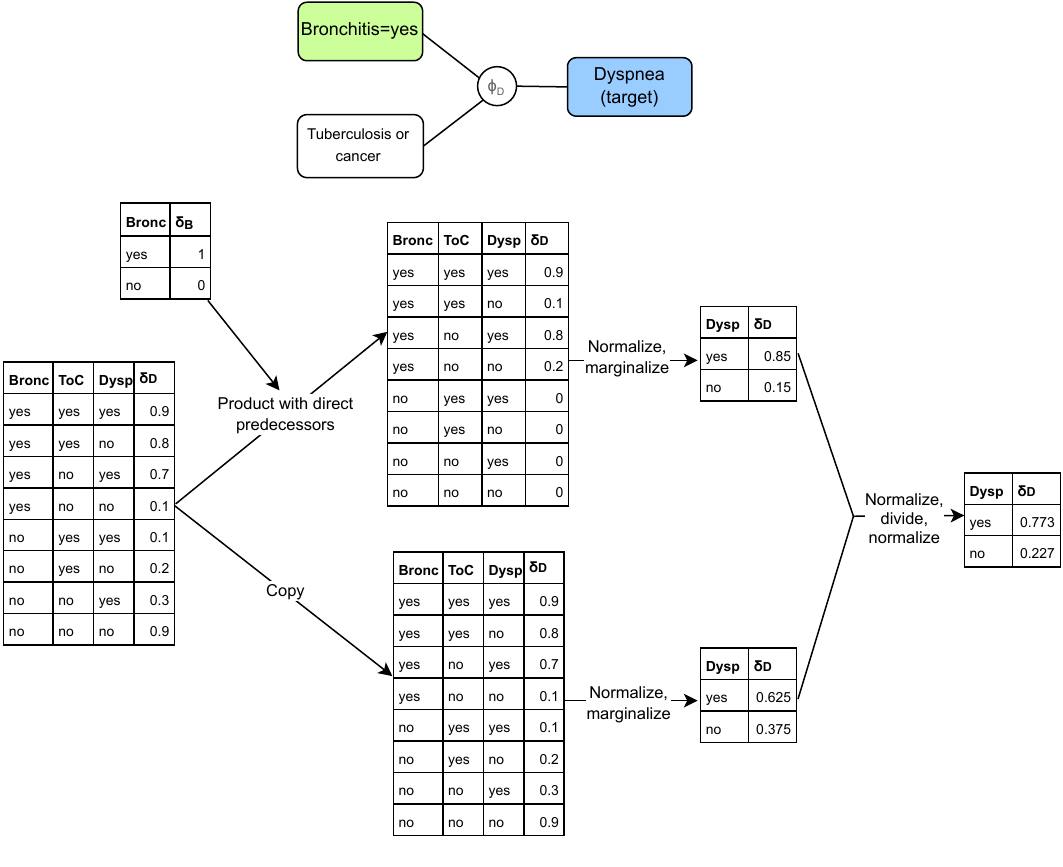}
    \caption{Visualization of Step Effect calculation. The numerator is calculated by multiplying the initial factor by incoming evidence factors, then normalizing, and marginalizing by the direct predecessor nodes. The denominator is calculated by marginalizing by the direct predecessor nodes and normalizing. After division, the resulting factor is normalized. Note that for compactness of visualization, we replaced explicit names of node states with ``yes'' and ``no'' notation. ``ToC'', ``Bronc'', and ``Dysp'' are the abbreviations of the corresponding nodes.}
    \label{fig:step_effect}
\end{figure}

An example of calculating a step effect is shown in Figure~\ref{fig:step_effect}.

% NOTE: The idea here is that according to the factor graph structure it is possible that one variable node can have several incoming connections from factor nodes. At the same time factor node can have several incoming connections from the variable nodes. Thus we need to find a way of precise definition of all the nodes from the factor graph from which we receive the signal. E.g. check the sample factor graph on this moment of the video https://www.youtube.com/watch?v=Yo-xat4cn8M&t=1441s 

The definition of \textit{SE} is reminiscent of the message-passing equations. However, we additionally perform division by the marginalized factor to separate the information derived from the updates $\Delta_{\phi}$ and the information passively contained in the factor $\phi$. Note that the definition of \textit{SE} is purely heuristic, and it does not correspond to any clear semantics. In section \ref{sec:proof-of-concept} we will later show that, in practice, this content determination algorithm approximates loopy message propagation.

Now, having defined the \textit{SE}, we may define the way to compute the factor argument effect (\textit{FAE}) updated beliefs of a complete \textit{FA} on each of the nodes it involves.

% This calculation involves recursively calculating the \textit{SE} along the \textit{FA} from the evidence nodes to its target node. If any node within the \textit{FA} is preceded by multiple factors, we multiply the \textit{SEs} of each direct predecessor factor. Formally, the \textit{FAE} of \textit{FA} on any node $X\in FA$ is calculated as shown in Eq.~\ref{eq:argument_effect}
% %  to compute the total FAE on node t

% % This calculation is performed by simply calculating the step effect along the FA from evidence nodes to a target node $t^*$, as shown in Eq.~\ref{eq:argument_effect}

% \begin{equation}
% \delta_{X} = FAE(FA, X) = \prod_{\phi \in Pred_{FA}(X)} SE(\phi, \Delta_\phi, X),
% \label{eq:argument_effect}
% \end{equation}

% NEW DEFINITION
\begin{definition}[Factor Argument Effect (\textit{FAE})]
Let $FA$ be a factor argument within a Bayesian Network, consisting of a sequence of factor nodes and variable nodes leading from evidence nodes to a target node. The Factor Argument Effect (FAE) of $FA$ on a target node $X \in FA$, denoted as $FAE(FA, X)$ or $\delta_{X}$, is defined as the cumulative effect of recursively calculating the $SE$ along the $FA$ from the evidence nodes to its target node $X$. Formally,

\begin{equation}
\delta_{X} = FAE(FA, X) = \prod_{\phi \in Pred_{FA}(X)} SE(\phi, \Delta_\phi, X),
\label{eq:argument_effect}
\end{equation}

where $\Delta_{\phi} = \{\delta_Y : Y \in \text{Pred}_{FA}(\phi)\}$ is a set of belief updates $\delta_Y$ for each node $Y$ preceding $\phi$ in the \textit{FA}. These belief updates are either computed recursively or are lopsided factors representing a piece of observed evidence. More formally,

\[
    \delta_Y= 
\begin{cases}
    \textit{FAE(FA, Y}), \text{if Y}  \notin \text{observed evidence} \\
    \textit{Obs(Y = y)},  \text{otherwise}
\end{cases},
\]

where \textit{Obs} is a function that returns a lopsided factor, and \textit{Obs(Y = y)} is a lopsided factor such as [$True$: $0$, $False$: 1] for the observation $Y=False$.

\end{definition}
% NEW DEFINITION ENDS

% More formally, $\delta_Y= FAE(FA, Y)$ if Y is not observed, otherwise $\delta_Y = \text{Obs(Y = y)}$. $\text{Obs(Y = y)}$ is a lopsided factor such as eg [$True$: $\epsilon$, $False$: 1] for the observation $Y=False$ (a small $\epsilon$ prevents division-by-zero).

 \begin{figure}[t!]
    \centering
     \includegraphics[width=\linewidth]{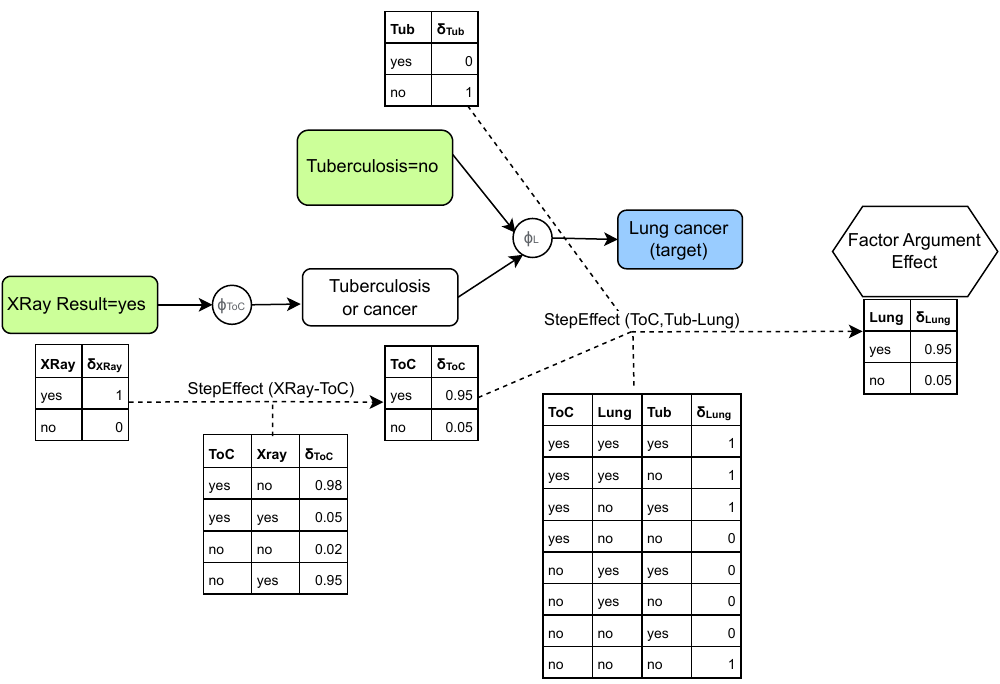}
    \caption{Visualization of Factor Argument Effect calculation. Step Effect is recursively calculated from evidence nodes to target node consequently updating the beliefs of all variable nodes within the Factor Argument. Note that for compactness of visualization, we replaced explicit names of node states with ``yes'' and ``no'' notation.``ToC'', ``Tub'', and ``Lung'' are the abbreviations of the corresponding nodes.}
    \label{fig:faeffect_intuit}
\end{figure}

 In Figure~\ref{fig:faeffect_intuit}, we show how to compute the \textit{FAE} by recursively applying \textit{SE}, consequently updating the beliefs of all variable nodes within the \textit{FA}. The definition of the \textit{FAE} allows quantifying the effect of \textit{FA} on our beliefs about each of its intermediate nodes. We will be able to build on this capacity later to produce textual explanations of \textit{FAs}.

\subsection{Factor argument strength}
\label{sec:arg_strength}

    The definition of the \textit{FAE} is a comprehensive description of how the \textit{FA} affects our beliefs of all nodes from evidence to target variable nodes. But if we want to only show to the user those \textit{FAs} that are most relevant, we need to define the way of comparing \textit{FA} effects to each other.  For this, we define the notion of \textit{factor argument strength} (\textit{FAS}) w.r.t. a value of the target variable $T = t_o$ as follows:
    
    % For this, we define the \textit{factor argument strength} (\textit{FAS}) w.r.t. a value of the target variable $T = t_o$ as follows:

    % \begin{equation}
    % FAS(FA, T=t_o) =
    % \log \frac{\delta_T(t_o)}{\frac{1}{N-1}\sum\limits_{i\neq o}\delta_T(t_i)}\label{eq:argument_strength}
    % \end{equation}
    % % https://gitlab.nl4xai.eu/jaime.sevilla/explainbn/-/blob/main/explainbn/arguments.py#L93 >> https://gitlab.nl4xai.eu/jaime.sevilla/explainbn/-/blob/main/explainbn/utilities.py#L41 

    % where $N$ is the number of possible states of the target variable $T$, $\delta_T = FAE(FA, T)$ is the effect of \textit{FA} on T, and $\delta_T(t)$ is the value assigned by the belief update $\delta_T$ to outcome $t$.

    % New definition
\begin{definition}[Factor Argument Strength (\textit{FAS})]
Given a Bayesian Network and a factor argument \(FA\) affecting the belief state of a target variable \(T\), the Factor Argument Strength (FAS) with respect to a particular value of \(T\) (denoted as \(T = t_o\)) is defined as follows:

\begin{equation}
    FAS(FA, T=t_o) =
    \log \frac{\delta_T(t_o)}{\frac{1}{N-1}\sum\limits_{i\neq o}\delta_T(t_i)}\label{eq:argument_strength}
\end{equation}

where:
\begin{itemize}
    \item \(N\) is the number of possible states of the target variable \(T\),
     \item \(\delta_T(t_o)\) is the Factor Argument Effect associated with the target variable taking the value \(t_o\) due to the factor argument \(FA\),
     \item \(\delta_T(t_i)\) is the Factor Argument Effect associated with all other states \(t_i\) of \(T\), excluding \(t_o\) due to the factor argument \(FA\).
\end{itemize}

\end{definition}
% Definition ends

    Our definition of \textit{FAS} satisfies one intuitive property: if the target node in the corresponding subgraph of the Bayesian Network associated with \textit{FA} is d-separated conditional on the evidence nodes of the FA, then its associated \textit{FAS} is zero. This can be seen from Eq.~\ref{eq:step_effect} -- if the reasoning step from the direct predecessors $\text{Pred}_{FA}(\phi)$ is d-blocked, then the equation simplifies to a constant factor, which propagates along its path. In Eq.~\ref{eq:argument_effect}, since all paths are d-blocked by assumption, the final node of all incoming \textit{SE} is constant. Finally, in Eq.~\ref{eq:argument_strength} the strength of a constant factor is 0.

    The idea behind \textit{FAS} is inspired by~\cite{doi:10.1080/10618600.1997.10474735}, where the likelihood ratio of the hypothesis given the evidence was used to quantify the strength of a chain of reasoning. We, however, use a notion of strength directly based on the CPTs of the BN, which allows us to talk about the strength of specific statistical relations rather than the strength of (a subset of) the evidence variables. We can use the \textit{FAS} to rank different \textit{FAs}, and select those that are most important to defend a given conclusion.

 \begin{figure}[t!]
    \centering
     \includegraphics[width=\linewidth]{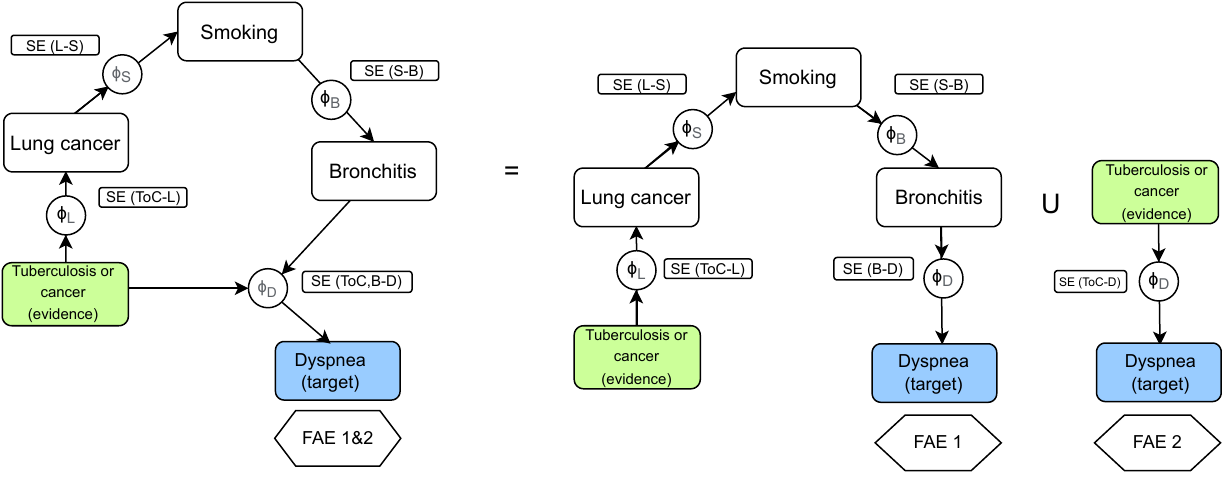}
    \caption{The complex factor argument (FA) can be decomposed into two simpler \textit{FAs}. This decomposition is a good approximation if the factor argument effect (FAE) of the complex argument is approximately similar to the product of the \textit{FAEs} of the simpler arguments. Arguments of \textit{SE} (ToC, L, S, etc) are the abbreviations of the corresponding nodes.}
    \label{fig:args_composition}
\end{figure}

\subsection{Splitting factor arguments}

During interaction with the BN, multiple pieces of evidence may be provided. Moreover, BNs may involve multiple simple paths between even a single evidence node and a target node. Consider an example in Figure~\ref{fig:args_composition}, where a single evidence node ``Tuberculosis or Cancer'' may be connected through two different simple paths to the target node ``Dyspnea''.

This fact raises the question whether these two \textit{FAs} should be delivered to the BN user jointly or separately. This problem was previously discussed, e.g., in \cite{keppens2012argument}, where the author explains the distinction between convergent arguments — where each of them independently supports a conclusion — and linked arguments — where the strength of each \textit{FA} depends on the presence of the other.

Thus, we need a formal way to decide whether the \textit{FAs} are independent. If they are, then we can break down the composite \textit{FA} into its simpler subcomponents. But if the interaction between the \textit{FAs} is important to support the conclusion, we need to present the more complex \textit{FA} to the user.  The formal definition of independent FAs is as follows:

% We define two \textit{FAs} as \textit{independent} if the product of the \textit{FAE}s of two separate \textit{FAs} is equal to the \textit{FAE} of the  \textit{FA} union, where \textit{FA} union refers to the union of corresponding graphs. More generally, we define a family of \textit{FAs} as independent if the product of the \textit{FAE}s of any subset of these \textit{FAs} is equal to the \textit{FAE} of the union of arguments in said subset.

% New definition
\begin{definition}[Independent Factor Arguments]
Two factor arguments (FAs) within a Bayesian Network are considered independent if, and only if, the product of the Factor Argument Effects (FAEs) of these separate FAs equals the FAE of their combined effect, where the combination of FAs refers to the union of their corresponding graphs. Formally, let $FA_1$ and $FA_2$ be two distinct FAs, and let $FA_{union}$ denote their union. The FAs $FA_1$ and $FA_2$ are independent if:

\begin{equation}
FAE(FA_1) \times FAE(FA_2) = FAE(FA_{union})
\end{equation}

\end{definition}
% New definition ends

As an illustration, we can consider what would happen in an AND network. This network has a $A\rightarrow C \leftarrow B$ structure where $A$ and $B$ nodes are distributed as a Bernoulli($1/2$), and $C$ is true \textit{iff} $A$ and $B$ are true. The \textit{FAE} of the argument $A=1$ on $C$ is $(C=1: 3, C=0:1)$, and similarly for $B=1$. However, the factor effect of the joint \textit{FA} $A=1, B=1$ on $C$ is $(C=1: 1, C=0:0)$. This is different from the product of the individual effects, so we conclude that any \textit{FA} we present to the user must present both \textit{FAs} jointly. See the implementation of this example in the demonstration notebook in our repository (see Section~\ref{sec:intro}).

% https://gitlab.nl4xai.eu/jaime.sevilla/explainbn/-/blob/main/explainbn/arguments.py#L193

% This definition captures the essence of independence we care about. 
In terms of explanations, the definition of independence is too restrictive - \textit{FAs} with weak interaction with one another should still be presented independently to the user. Thus, we define the notion of approximate independence of \textit{FAs} as follows:

% Thus, we define the notion of \textit{approximate independence} of \textit{FAs}, which states that a set of \textit{FAs} is approximately independent if the product of the \textit{FAE}s of any subset of the \textit{FAs} is within a certain threshold of distance from the \textit{FAE} of the union of \textit{FAs} in said subset.

\begin{definition}[Approximate Independence of Factor Arguments]
A set of Factor Arguments (FAs) within a Bayesian Network is said to be approximately independent if the product of the Factor Argument Effects (FAEs) of any subset of these FAs is within a specified threshold of distance from the FAE of the union of FAs in said subset. Formally, for a given threshold $\theta$, a set of FAs $\{FA_1, FA_2, \ldots, FA_n\}$ is approximately independent if for any non-empty subset $\{FA_i\} \subseteq \{FA_1, FA_2, \ldots, FA_n\}$, the following condition holds:
\begin{equation}
\left| FAE(\bigcup_{FA_i} FA_i) - \prod_{FA_i} FAE(FA_i) \right| < \theta,
\end{equation}
where $\bigcup_{FA_i} FA_i$ denotes the union of the factor arguments in the subset, and $FAE(FA_i)$ represents the Factor Argument Effect of $FA_i$.

\end{definition}

This notion captures the degree to which the combined effect of a set of FAs deviates from what would be expected if they were truly independent, based on their individual effects.

To decide whether the effect of the union of arguments is close to the product of the effects of the individual arguments we introduce the notion of factor argument distance (\textit{FAD}). The formal definition of measuring the distance between two \textit{FAE}s combines the ideas of \textit{FAE} and \textit{FAS}. The \textit{FAE} inferred with a certain \textit{FA} about the target variable $\delta_{T}$ is expressed as a factor. To quantify the difference between two $\delta_T$ factors, we may divide one factor by another and then look for the biggest resulting \textit{FAE} from all possible states of the target within the divided factor. The formal definition of \textit{FAD} is as follows:

% New definition
\begin{definition}[Factor Argument Distance (\textit{FAD})]
The Factor Argument Distance (FAD) between two factor arguments (FAs) within a Bayesian Network, considering their effects on the belief state of a target variable $T$, is defined through the differential impact of these FAs, quantified by their FAEs. This is captured by the equation:

\begin{equation}
\begin{split}
FAD(\delta_{T}^a, \delta_{T}^b) =
\max_{i < N} | \log \frac{ \delta_{T}^{a/b}(t_i)}{\frac{1}{{N-1}}\sum\limits_{j\neq i}\delta_{T}^{a/b}(t_j)} | \label{eq:factor_distance}
\end{split} 
\end{equation}

where $\delta_T^a, \delta_T^b$ are two different belief updates, $\delta_{T}^{a/b}(t_i)$ is a shorthand for $\delta_{T}^a(t_i)/ \delta_{T}^b(t_i)$ and we assume that variable $T$ has $N$ possible outcomes numbered from $0$ to $N-1$.
\end{definition}
% Definitio ends
% https://gitlab.nl4xai.eu/jaime.sevilla/explainbn/-/blob/main/explainbn/utilities.py#L100

Intuitively, the definition of \textit{FAD} corresponds to identifying the outcome $t_i$ on which factors $\delta_T^a, \delta_T^b$ disagree the most, and then quantifying in terms of logodds the difference between the respective updates they would induce if applied to variable $T$.

This definition yields small values if the \textit{FAE} about the target variable $\delta_{T}$ inferred by two different \textit{FAs} are similar, and bigger values otherwise. This is fairly easy to see, since if $\delta_T^a \approx \delta_T^b$ then we will have that $\delta_{T}^{a/b}(t_i) \approx 1$, and so $FAD(\delta_{T}^a, \delta_{T}^b) =
\max_{i < N} | \log \frac{ \delta_{T}^{a/b}(t_i)}{\frac{1}{{N-1}}\sum\limits_{j\neq i}\delta_{T}^{a/b}(t_j)} | \approx \max_{i < N} | \log \frac{ 1}{\frac{1}{{N-1}}\sum\limits_{j\neq i}1} | = 0$. For dissimilar factors, the FAD will be above zero. As desired, we can use this to decide if the effect of the union of arguments is close to the product of the effects of the individual arguments.  

We finally define a \textit{FA} to be approximately proper:

% if it cannot be expressed as the union of (approximately) independent \textit{FAs}. 

\begin{definition}[Approximately Proper Factor Argument]
A Factor Argument (FA) within a Bayesian Network is deemed approximately proper if it cannot be decomposed into a union of approximately independent FAs. This condition implies that the FA, as a whole, contributes uniquely to the inference process without being reducible to simpler, approximately independent components. Formally, an FA is approximately proper if, for any partition of the FA into subsets $\{FA_1, FA_2, \ldots, FA_n\}$, where each $FA_i$ is an approximately independent factor argument, there exists no combination of these $FA_i$'s whose union effectively replicates the inference impact of the original FA within a specified approximation threshold.

\end{definition}

This definition ensures that an \textit{FA} is considered approximately proper only if its contribution to the Bayesian inference cannot be approximated by combining the effects of smaller, simpler arguments, thereby maintaining the integrity and the unique inferential value of the \textit{FA} within the network's reasoning process.

Our next goal will be to identify all the proper \textit{FAs}  relating the available evidence to a target. We explain how to do this in the next section.

\begin{algorithm*}[t!]
\small
  \caption{Overview of the content determination algorithm for Bayesian Network explanation}\label{algo}
  \begin{algorithmic}[]
    \Function{FindMaximalProperFAs}{$BN, t, E$} 
        
        \State \textbf{Input}: $BN$ \Comment{Bayesian Network in factor view}
        \State \textbf{Input}: $t$ \Comment{target node}
        \State \textbf{Input}: $E$ \Comment{evidence nodes}
        \State \textbf{Output}: $ProperAPs$ \Comment{list of independent factor arguments}
        \State \textbf{Constant} $ML$ \Comment{max length of the simple path}
        \State \textbf{Constant} $MC$ \Comment{max number of simple paths}
        \State \textbf{Constant} $DT$  \Comment{Dependence Threshold}
        \State AllSimplePaths = GetAllSimplePaths(BN, t, E, ML)
        % \Comment{Paths from evidence nodes to target node}

        \For{$SPCombination \in Combinations(AllSimplePaths, MC)$}
            \State IsDependent = CheckDependence(SPCombination,DT)
            \If{IsDependent}
                \State Add ComposeFAs(SPCombination) to ProperFAs
            \EndIf 
        \EndFor

        \State ProperFAs = AdjustProperFAs(ProperFAs)

        \State OutputFAs = FilterNonMaximalFAs(ProperFAs) 
        % \Comment{Remove non-maximal arguments}

        \If {ML $<$ $\infty$ or MC $<$ $\infty$}
            \State OutputFAs = PairwiseCombine(OutputFAs) 
            \State  \Comment{Guarantees pairwise independence when heuristics are applied}
        \EndIf
    
        \Return OutputFAs
  
    \EndFunction
    
    \Function{CheckDependence}{$SPCombination, DT$}
        
        \State SPCombinationUnion = ComposeFAs(SPCombination)
        \State SPCombinationUnionEffect = FAEffect(SPCombinationUnion) \Comment{Eq.~\ref{eq:argument_effect}}
        \For{$SPCombinationUnionPart \in Partitions(SPCombination)$}
            \State SPCombinationUnionPartEffect = 
            \State \hspace{12em} FAEffect(SPCombinationUnionPart)
            \State  Distance = FADistance(SPCombinationUnionEffect,                 \State  \hspace{12em} SPCombinationUnionPartEffect) 
            \Comment{Eq.~\ref{eq:factor_distance}}
            \If{Distance $<$ DT} 
                \Return False
            \EndIf
        \EndFor
        
        \Return True

    \EndFunction

    \Function{AdjustProperAPs}{$ProperAPs$}

        \State ProperAPs = RemoveSubgraphs(ProperAPs)
        % \Comment{Remove \textit{FA} which is a subgraph of any other FA in ProperAPs}        
        \State ProperAPs = PairwiseRefinement(ProperAPs)
        \State \Comment{Combine non-independent pairs of FA until all FA are independent}       

    \Return ProperAPs

    \EndFunction    

    % \Function{RemoveSubarguments}{$ProperAPs$}

    % \For {$AP_i \in ProperAPs$}
    %     \For {$AP_j \in ProperAPs$}
    %         \If {$AP_i.isSubgraph(AP_j)$}
    %             \State $min \leftarrow e(v_i, v_j) + l(v_j)$
    %             \State $p(i) \leftarrow v_j$
    %         \EndIf
    %     \EndFor
    %     \State $l’(i) \leftarrow min$
    % \EndFor

    % \EndFunction    
    
  \end{algorithmic}
\label{algo:algo}
\end{algorithm*}

\subsection{Core algorithm overview}

At this stage, we are ready to define our content determination approach for the BN reasoning explanation. Its idea is to construct a set of proper, maximal, and independent \textit{FAs} relating the evidence to the target node. Note that maximal \textit{FAs} are defined as \textit{FAs} that are not a subgraph of another \textit{FA}. The pseudocode of the algorithm is shown in Algorithm~\ref{algo:algo}.

The inputs to the algorithm are the factorized BN, the target node, and the observed evidence. First, we construct all \textit{FAs} corresponding to simple paths from evidence nodes to the target node. Each possible combination of simple paths passes the dependence check, which verifies whether the corresponding composed \textit{FA} is proper in the \textit{CheckDependence} function (note that every FA can be expressed as a combination of simple paths, so this loop exhausts all FAs). To calculate this, we compare the \textit{FAE} of the composed \textit{FA} to the product of the \textit{FAE} of the union of each possible partition of the arguments that compose it using \textit{FAD} between these \textit{FAE}. If the \textit{FAD} between the \textit{FAE} of any possible partition and the \textit{FAE} of the composed \textit{FA} is below the given threshold  (parameter \textit{DT}), then we conclude that the current combination of \textit{FAs} is independent; hence, it is not to be delivered jointly. Otherwise, if the \textit{FAD} of all possible partitions is always above the threshold, we conclude that the composed \textit{FA} is proper and will be delivered jointly. Finally, we filter for maximal \textit{FAs}. The result is a set of maximal, proper, and independent \textit{FAs} that capture how the evidence relates to the target. 

Note that we have omitted two optimizations in the \textit{CheckDependence} function for simplicity of exposition. First, we only check for partitions composed of \textit{FAs} previously identified to be proper. Second, the \textit{FAEs} of each composed argument are stored and dynamically reused between calls. To facilitate these optimizations, we need to iterate from \textit{SPCombinations} of fewer \textit{FAs} to those of more \textit{FAs}. The implementation details are available in our repository (see Section~\ref{sec:intro}).

% Several problems may occur after the list of proper FAs is collected, and they are addressed in the AdjustProperAPs function which sequentially launches two following functions. The RemoveSubgraphs function drops such proper FAs that turn out to be the subgraph of more complex FAs. The PairwiseRefinement function makes sure that none of the FAs are mutually dependent, which can sometimes be the case because of the limitation of \textit{FA} complexity

The result of the proposed algorithm is the list of \textit{FAs} that will be delivered to the user, separately and in descending order w.r.t. their \textit{FAS} (Eq.~\ref{eq:argument_strength}). Typically, we show only a handful of these \textit{FAs} - namely those that have the highest strength and thus are most relevant for the user. We can control how many \textit{FAs} to show by filtering those with \textit{FAS} below a threshold and/or only showing the top N \textit{FAs}. Once we have selected the \textit{FAs} to show to the user, we need to explain them in natural language. 
%We explain how to do this in the next section.

\subsection{Ways to overcome algorithm limitations}

The proposed algorithm has certain limitations. Since the number of simple paths in a graph is factorial to the number of variables and the number of complex \textit{FAs} is exponential on the number of simple paths, the naive approach of listing all \textit{FAs} and computing all effects will be impractical for big networks. Some heuristics can relieve these issues. First, we can consider only simple paths with lengths below a threshold (parameter \textit{ML}). Second, we can only consider complex \textit{FAs} combined from a limited number of simple \textit{FAs} (parameter \textit{MC}).

These heuristics void the guarantee that the output will be a set of independent \textit{FAs} since some \textit{FAs} combinations will not be tried. To alleviate this issue, when applying the heuristics, we iteratively combine pairs of dependent \textit{FAs} to guarantee pairwise independence. This procedure removes the guarantee that the constructed \textit{FAs} will be proper. The last steps to verify that the selected \textit{FAs} are proper are to verify that none of these \textit{FAs} is a subgraph of another and to make sure that none of the \textit{FAs} are mutually dependent. 

\subsection{Textual explanation with extracted factor arguments}
\label{sec:text_expl}

Having identified the set of \textit{FAs} to be presented, the final step is explaining these \textit{FAs} in natural language. Refer to Table~\ref{tab:text_explanations} for the examples of the different types of the explanation. The details of these explanation types are described below. 
% We propose a template-based approach, which varies relatively to the reasoning patterns (causal, evidential, and intercausal)~\cite{koller2009probabilistic}.

\begin{table}[!t]
\centering
\begin{tabularx}{\textwidth}{lX}
 \hline 
\textbf{Type} & \textbf{Text of explanation}   \\
 \hline 
Overview         & Since $<$XRay Result$>$ is $<$abnormal$>$ and $<$Tuberculosis$>$ is $<$absent$>$, we infer that $<$Lung Cancer$>$ = $<$present$>$. \\ \hline
Direct           & We have observed that $<$XRay Result$>$ is $<$abnormal$>$ and $<$Tuberculosis$>$ is $<$absent$>$.

The updated probability of $<$XRay Result$>$ = $<$abnormal$>$ is evidence that the intermediate node $<$Tuberculosis or Cancer$>$  becomes strongly more likely to be $<$true$>$.

The updated probability of $<$Tuberculosis or Cancer$>$ = $<$true$>$ and $<$Tuberculosis$>$ = $<$absent$>$ is evidence that the target node   $<$Lung Cancer$>$ becomes strongly more likely to be $<$present$>$.  \\ \hline                                                     
Contrastive      & We have observed that $<$XRay Result$>$ is $<$abnormal$>$ and $<$Tuberculosis$>$ is $<$absent$>$.

   The updated probability of $<$XRay Result$>$ = $<$abnormal$>$ is evidence that the intermediate node $<$Tuberculosis or Cancer$>$  becomes strongly more likely to be $<$true$>$.
   
  Usually, if the $<$Tuberculosis or Cancer$>$ = $<$true$>$ then the $<$Lung Cancer$>$ = $<$true$>$.      
  
  Since the $<$Tuberculosis$>$ is $<$absent$>$, we can be strongly more certain that $<$Lung Cancer$>$ = $<$true$>$. \\ \hline
\end{tabularx}%
\caption{Three types of converting an \textit{FA} from Figure~\ref{fig:reasoning_sample} into textual form.}
\label{tab:text_explanations}
\end{table}%

\subsubsection{Explanation of Factor Argument steps}

Overall, the explanation of \textit{FA} is aimed at verbally guiding the BN user from the evidence node or nodes to the target node using the combination of the templates. The explanation of each step within \textit{FA} could include the explanation of observations, inference rules, and conclusion. 

Each \textit{FA} starts with a description of the evidence nodes. The description of the evidence is performed using the following template: \textit{We have observed that \{evidence node\} is \{evidence node state\}} (e.g., ``We have observed that $<$XRay Result$>$ is $<$abnormal$>$'').

After the evidence nodes are verbalized, all next steps until the target node include the description of premises that describe the updated beliefs of the nodes w.r.t. the previous \textit{FA} step, inference rules applied on the current \textit{FA} step, and the conclusion from the application of the rules.

The premises are described using the following template: \textit{The updated probability of \{previous node in FA\} = \{state of previous node in FA \}} (e.g., ``The updated probability of $<$XRay Result$>$ = $<$abnormal$>$''). The exact state of the node is selected to be verbalized if it is the state of the evidence node, or if this state has the highest magnitude of the logodds update.

The description of inference rules and their conclusions slightly varies according to the reasoning patterns applied at a certain step within \textit{FA}: evidential, causal, or intercausal~\cite{koller2009probabilistic}.

If the reasoning pattern of the particular step within \textit{FA} corresponds to causal or evidential reasoning, we use the following template, which will be further referred to as direct explanation: \textit{\{premises\} \{verb\} \{description of the conclusion with a strength qualifier\}} (e.g., ``The updated probability of $<$XRay Result$>$ = $<$abnormal$>$ is evidence that the intermediate node $<$Tuberculosis or Cancer$>$  becomes strongly more likely to be $<$true$>$'').

The verb is either \textit{``causes''} or \textit{``is evidence that''} for causal and evidential reasoning, respectively. The strength qualifier reflects the magnitude of beliefs logodds change according to the following mapping: \textit{``Certainly''}, \textit{``Strongly''}, \textit{``Moderately''}, \textit{``Weakly''}, and \textit{``Tenuously''} for cases when the range of logodds update is greater than 10, 1, 0.5, 0.1, and less than 0.1, respectively.

If the reasoning pattern of the particular step within \textit{FA} corresponds to intercausal there could be two options of explanation: direct and contrastive. In the direct explanation, the description is the same as in the purely evidential case we described above. In the contrastive explanation, we compute the counterfactual effect we would have gotten had we observed the premise corresponding to the child of the conclusion node, but not the co-parents (the node or nodes indirectly connected through a common child node with a target node), and we compare that to the actual inference.

The template of the contrastive explanation consists of two parts. The first part is: \textit{Usually, if \{child potential premises\} then \{counterfactual outcome description\}} (e.g., ``Usually, if the $<$Tuberculosis or Cancer$>$ = $<$true$>$ then the $<$Lung Cancer$>$ = $<$true$>$'').

It describes the effect we would have gotten had we observed the premise corresponding to the most likely state of the conclusion node's child, where the state of the child node (child potential premises) and the state of the target node within current \textit{FA} step (counterfactual outcome description), which become more likely within the \textit{FA} are described in plain text. 

The second part of the template is \textit{Since \{co-parent's premises\}, \{factual outcome description\} } (e.g., ``Since the $<$Tuberculosis$>$ is $<$absent$>$, we can be strongly more certain that $<$Lung Cancer$>$ = $<$true$>$'').

The co-parent's premises are verbalized in plain text (similar to the first part of the template), while the factual outcome description varies according to the results of the counterfactual outcome. If the most likely state of the target node corresponding to the counterfactual outcome contradicts the state inferred from the actual outcome of the \textit{FA} step, then the factual description is verbalized as \textit{we infer \{conclusion description\} instead}, where the conclusion description is just a plain text description of the most likely state of the target node within current \textit{FA} step. If the counterfactual outcome corresponds to the actual outcome, we use the template \textit{we \{verb\} be more certain that \{actual outcome description\}}, where verb \textit{``can''} is used if logodds of the most likely state of target node inferred from actual inference are higher than those inferred from counterfactual inference. Otherwise, if the actual inference logodds are not greater than the counterfactual ones, the verb \textit{``can not''} is used. 

Overall both parts of the explanation could be as follows: ``Usually, if the $<$Tuberculosis or Cancer$>$ = $<$true$>$ then the $<$Lung Cancer$>$ = $<$true$>$. Since the $<$Tuberculosis$>$ is $<$absent$>$, we can be strongly more certain that $<$Lung Cancer$>$ = $<$true$>$'').

Our approach implies that the \textit{FA} may contain multiple simple paths; thus, it is necessary to define the way to convert them into text. We opt for chaining together the explanation of each rule inferred by the factor node and then adding an extra line explaining the cumulative effect of all the rules (e.g., ``The updated probability of $<$XRay Result$>$ = $<$abnormal$>$ is evidence that the intermediate node $<$Tuberculosis or Cancer$>$ becomes moderately more likely to be $<$true$>$. The updated probability of $<$Dyspnea $>$ = $<$present$>$ is evidence that the intermediate node $<$Tuberculosis or Cancer$>$  becomes moderately more likely to be $<$true$>$. All in all, the intermediate node $<$Tuberculosis or Cancer$>$ becomes strongly more likely to be $<$true$>$'').

\subsubsection{Explanation of the whole Factor Argument}

We define three modes of explanation of the whole \textit{FA}: direct, contrastive, and overview. The direct mode verbalizes all \textit{FA} evidence and intermediate steps using the direct template defined above. The explanation in contrastive mode is equal to direct except for the intercausal reasoning \textit{FA} steps, where the contrastive template explanation is used. Finally, the overview produces a simple description of the observations and the conclusion of the \textit{FA} without verbalizing any intermediate steps within the \textit{FA}. Refer to the Table~\ref{tab:text_explanations} for all three types of explanations corresponding to the \textit{FA} from Figure~\ref{fig:reasoning_sample}.

\section{Experimental results}
\label{sec:proof-of-concept}
    Our definition of \textit{FAE} is, in a loose sense, meant to imitate the loopy message-passing inference algorithm. We can empirically check the quality of the approximation by comparing beliefs about certain target nodes given certain evidence node values calculated by the message-passing algorithm and our approach.

    To perform this comparison, we need to define the way of calculating the belief given the prior belief and \textit{FAE}. Let $\mathcal{FA}$ be the set of relevant, independent \textit{FAs} found by our algorithm w.r.t. the evidence nodes $E$ and target nodes $t$. The posterior is calculated as follows: $$\hat O(t |\mathcal{FA}) =  O(t) \cdot \prod_{FA\in \mathcal{FA}} FAE(FA, t)$$ where $O(t)$ is a factor representing the prior probability of the target variable as computed by the message passing algorithm. This equation, namely, defines the way of approximating the posterior probability of the target node as the product of the prior probability and the \textit{FAE}s of each relevant \textit{FA} on the target.
    
    To study the quality of the approximation, we take BNs of different sizes (from 5 to 37 nodes), collected from the bnlearn website~\cite{bnlearn}, and perform 200 iterations of comparison. Each iteration includes a random choice (with a random seed corresponding to the iteration number) of the evidence nodes and target nodes and further comparison of the probabilities inferred by our algorithm and message-passing algorithm. The main intuition of the definite BN selection was to start the experiments with the smallest BNs and gradually increase the size and treewidth of the BN until the computation time per iteration starts being unreasonable. This selection approach yielded seven BNs with a maximum size of 37 nodes and treewidth equal to 4.

\begin{figure}[t!]
    \centering
    \includegraphics[width=\linewidth]{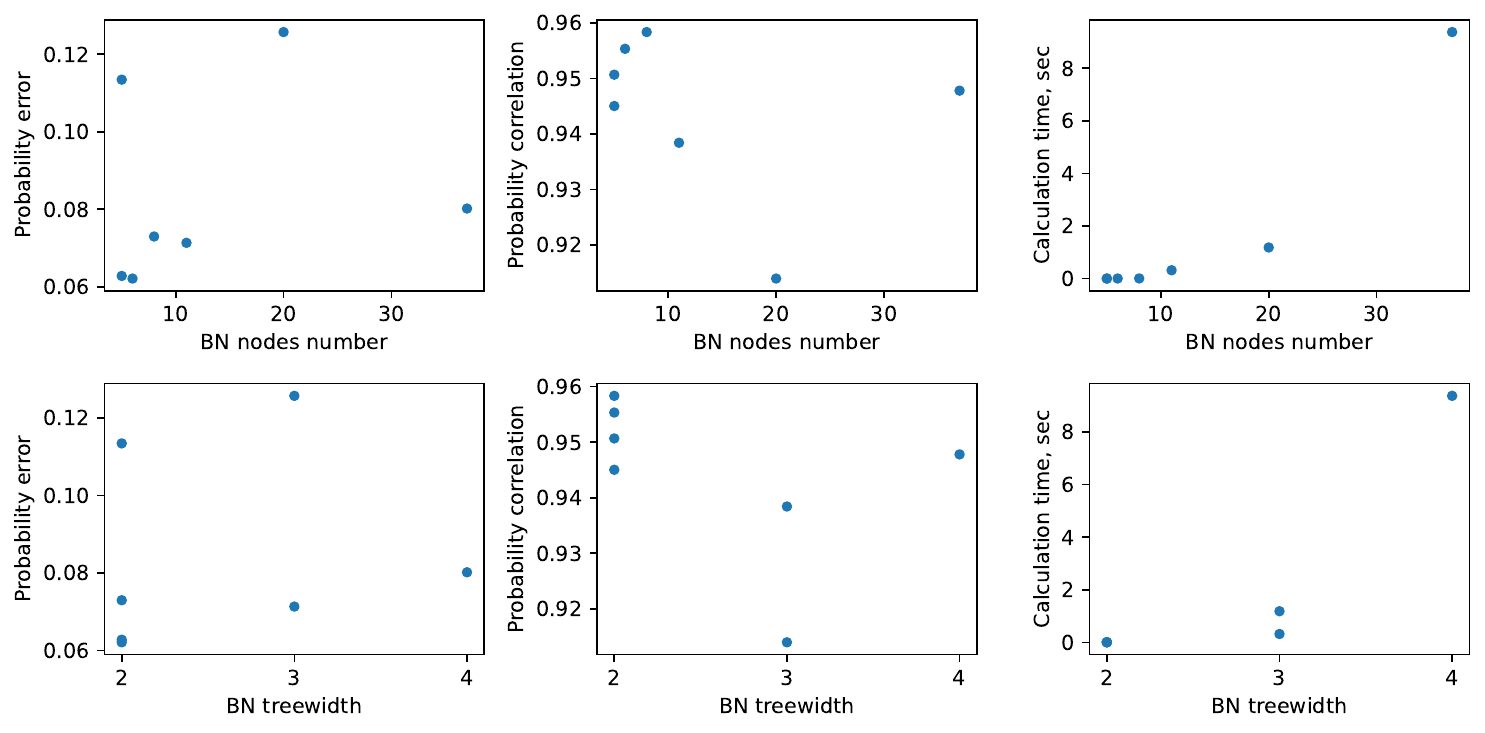}
    \caption{Mean absolute probability error of our approximation vs message passing, Spearman rho correlation coefficient, and calculation time from 200 runs of our algorithm with random target and evidence nodes with different BNs from bnlearn website (cancer, earthquake, survey, asia, sachs, child, alarm). We express the relations both in terms of the total BN node number and the treewidth. The calculations are performed with MC = 2.}
    \label{fig:run_different_bns_stat}
\end{figure}

We show the results of our experiments in Figure~\ref{fig:run_different_bns_stat} \footnote{The calculations for this figure are performed with an Intel(R) Core(TM) i7-8565U CPU @ 1.80GHz}, where we report the mean probability error and correlation between message passing and our algorithm, as well as the average computational time. We report these values w.r.t. the BN nodes number (because this may give an initial intuition about the size of the BN) and also w.r.t. the treewidth because it is known that the complexity of BN inference grows with the treewidth of the BN's graph~\cite{darwiche2009modeling}. To make the calculation faster, we limited the maximal complexity (MC parameter) of the \textit{FA} up to 2. The mean error between probabilities inferred by message passing and our algorithm is somewhat large, varying from 0.07 to 0.14. However, the Spearman correlation between these probabilities is rather high: 0.92 and above (the p-value for all of these cases was significantly less than 0.05). Both, mean error and correlation, do not have a visible trend w.r.t. the increase of either BN nodes number or treewidth. This suggests that the \textit{FAE} approximation of the message-passing process is qualitatively correct.

% \begin{table}[t!]
% \centering
% % \small
% \begin{tabular}{lccrrrr}
% \hline
% \textbf{BN} &  \textbf{\#nodes } &  \textbf{density} & $\mu_{err}$ &  $\rho_{prob}$ &    \textbf{sec/iter }\\
% \hline
%     cancer &        5 &     0.20 &         0.06$\pm$0.01 &  0.94 &      0.01 \\
% earth &        5 &     0.20 &         0.09$\pm$0.02  &  0.94 &      0.01 \\
%     survey &        6 &     0.20 &         0.07$\pm$0.01  &  0.94 &      0.02 \\
%       asia &        8 &     0.14 &         0.07$\pm$0.01  &  0.96 &      0.02 \\
%      sachs &       11 &     0.15 &         0.14$\pm$0.02  &  0.89 &      1.00 \\
%      child &       20 &     0.07 &         0.12$\pm$0.01  &  0.92 &      2.07 \\
%      alarm &       37 &     0.03 &         0.08$\pm$0.02  &  0.96 &      42.58 \\
% \hline
% \end{tabular}
% \caption{The statistics from 200 runs of our algorithm with random target and evidence nodes with different BNs and a maximum number of simple \textit{FAs} in a complex \textit{FA} equal 2. $mu_{err}$ and $\rho_{prob}$ stand for mean absolute error and Spearman correlation between posterior probabilities inferred by message passing and our algorithm, respectively. The BNs are taken from bnlearn.com.}
% \label{fig:run_different_bn}
% \end{table}

\begin{figure}[t!]
    \centering
    \includegraphics[width=\linewidth]{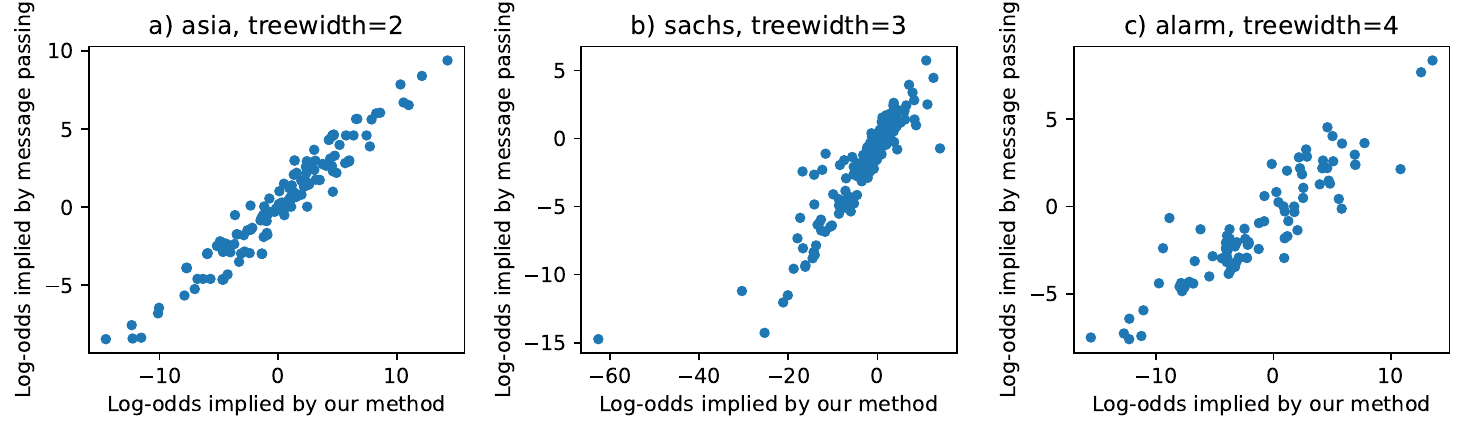}
    \caption{Correlation between the implied log-odds of the query according to our algorithm (x-axis) vs message passing (y-axis). We show the results for three BNs with different treewidth. The calculations are performed with MC = 2.}
    \label{fig:corr}
\end{figure}

Figure~\ref{fig:corr} further illustrates the correlation between the two methods. The logodds implied by our algorithm and the logodds computed through message passing are tightly correlated. However, the slope of the correlation is less than 1, suggesting that our algorithm overweight the strength of the argument. Further work could look into understanding why this is, and how to adjust the strength of the arguments to correct it. Nevertheless, we think that these results suggest that the algorithm is qualitatively focusing on the right parts of the explanation.

\begin{figure}[t!]
    \centering
    \includegraphics[scale=0.65]{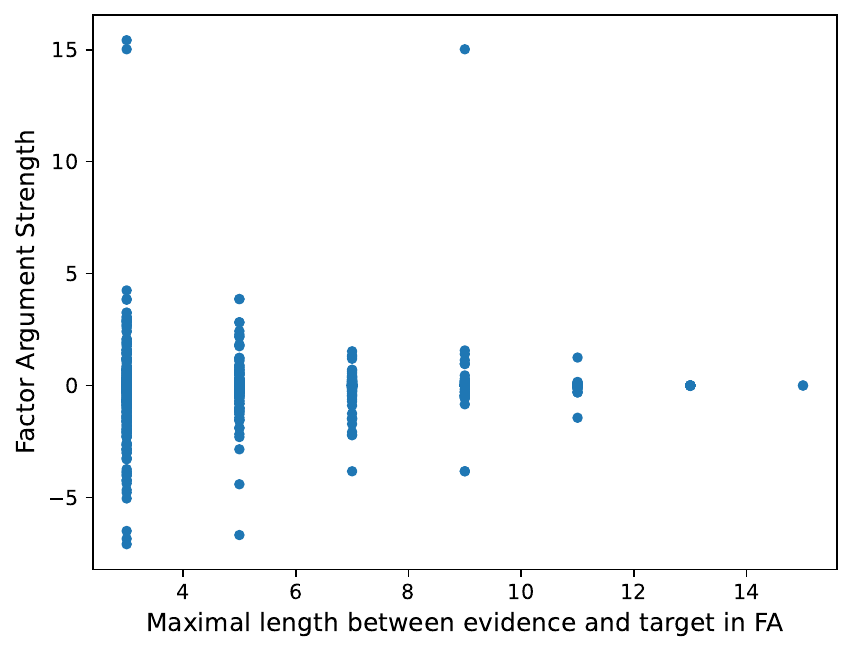}
    \caption{The statistics of proper \textit{FAs} length and \textit{FA} strength from 200 runs of our algorithm with random target and evidence nodes with different BNs from bnlearn website.  The calculations are performed with MC=2.}
    \label{fig:fa_lengths_vs_fa_strength}
\end{figure}

However, Figure~\ref{fig:run_different_bns_stat} also shows the main drawback of the approach: even limiting the complexity of the \textit{FAs} considered, the processing time increases significantly with the increase in the BN nodes number and its treewidth. Thus, we conclude that the proposed approach may rather precisely imitate the exact inference of the message passing algorithm; however, it is rather to be used for small or medium-sized BNs (around 20 nodes and treewidth equal to 2), and using it with bigger BNs requires further optimization of the algorithm.

We also collect the statistics of certain properties of the proper \textit{FAs} generated with our algorithm. In particular, we check whether the shorter \textit{FAs} are more often resulted into proper ones. To check this, similarly to the previous experiments, we run \textit{FA} calculations on the BNs from bnlearn website with random evidence and target nodes and plot the statistics of \textit{FA} length and corresponding \textit{FA} strength. The \textit{FA} length is calculated as a maximal path between evidence and target nodes with given \textit{FA}. Figure~\ref{fig:fa_lengths_vs_fa_strength} confirms the intuition that the shorter \textit{FAs} are more often considered proper by our algorithm and also tend to have greater \textit{FAS} than longer \textit{FAs}.

\begin{figure}[t!]
    \centering
    \includegraphics[width=\linewidth]{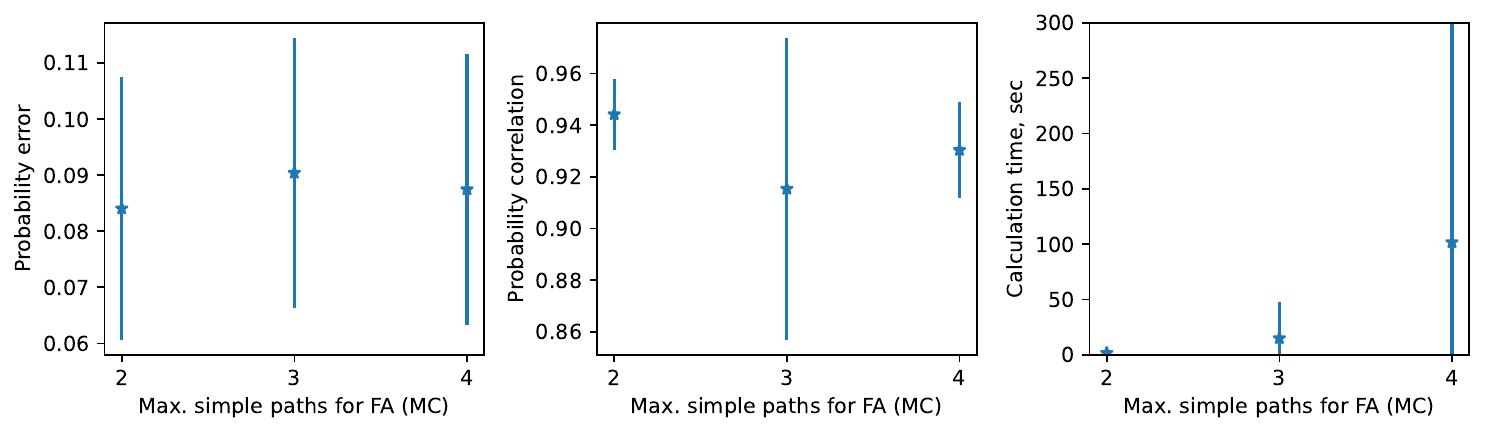}
    \caption{The statistics from 200 runs of our algorithm with random target and evidence nodes with different BNs from bnlearn website related to different values of the MC parameter (the maximum complexity of the resulting \textit{FA}).}
    \label{fig:mc_different_metrics}
\end{figure}

% To make sure that the number of nodes is the main factor affecting the computation time, we also report the density of the BNs. Higher values of density correspond to more edges; however, in the reported values, we can clearly see that the density decreases while the number of nodes increases, which confirms the intuition about the number of nodes being the main reason for the computational time increase. The code for the reproduction of the computational experiment is available in the Supplementary Materials.
% %, which is calculated as $density=E/(N(N-1))$, where E is the number of edges and N is the number of nodes of the BN.

Whereas limiting the complexity of \textit{FAs} may be used as a heuristic for decreasing calculation time, it is necessary to make sure that the \textit{FAs} generated under these heuristics are still qualitatively correct. As far as we have seen from Figure~\ref{fig:fa_lengths_vs_fa_strength} our algorithm is prone to generating shorter \textit{FAs} it seems that the heuristic of limiting maximal complexity of the presented \textit{FAs} (MC parameter) is more applicable than limiting the maximal length of \textit{FAs} (ML parameter). Thus we study the dynamic of error and correlation of posterior probabilities between our algorithm and message passing algorithm, as well as the computational time with the corresponding standard deviation w.r.t. the maximal complexity of \textit{FA} (MC parameter). Figure~\ref{fig:mc_different_metrics} shows that there are no significant differences between error and correlation within different values of \textit{FA} complexity. At the same time, the computational time increases significantly with the increase of MC. Thus it seems practical to use MC equal to 2 for practical usage of the proposed algorithm because it yields reasonable probability error and computation time. 
% , which implies that in the current state, the algorithm may be practically used only with small MC.

\section{Human-driven evaluation of the explanation method in the medical domain}
\label{sec:human_eval}

In this section, we compare our algorithm with two alternative algorithms using human-driven evaluation relying on widely-known BN describing lung diseases, as well as some of their potential causes and consequences, which was originally composed for demonstration purposes (see Figure~\ref{fig:reasoning_sample}). 

\subsection{BN explanation methods for comparison}

The first compared method, referred to as a \textit{baseline} simply verbally describes how the probability of the target node is updated given the provided evidence. The textual information is accompanied by graphical tips that include two screenshots of the BN from Netica software\cite{netica} before and after the evidence is provided. This idea is similar to the parts of evaluation pipelines in~\cite{kyrimi2020incremental,BUTZ2022102438} where simply showing BN without clarification was used among other explanation approaches.

The alternative textual explanation approach we perform the comparison to, referred to as \textit{incremental}, was described in~\cite{kyrimi2020incremental}. Its main idea is to distinguish the evidence that is considered important between supporting and non-supporting the final change in the target variable. The explanations are accompanied by Netica screenshots highlighting the nodes mentioned in the explanation. We find this method the most suitable for direct comparison because, to the best of our knowledge, this is the only previously proposed algorithm that can be used for textual explanations of any type of BN inference with the ability to set arbitrary nodes both as target and as evidence nodes.

Our method is referred to as \textit{fae}. It also delivers a textual explanation using direct explanation mode together with a visual aid in the form of Netica screenshots with the additional arrays of the \textit{FAs}' direction. Examples of the interface of all explanation methods used for human evaluation can be found in Figures~\ref{fig:baseline_example},\ref{fig:increm_example},\ref{fig:increm_example_23},\ref{fig:fse_example} in~\ref{sec:appendix_explanation}.

\subsection{Evaluation setup}

In general, the human-driven evaluation of BN reasoning explanations is a complex task. To the best of our knowledge, there is currently only one study explicitly dedicated to this task~\cite{BUTZ2022102438}. Moreover, the novel explanation methods, when presented, are not generally compared to the existing ones or demonstrated to the potential users of the explanation~\cite{TIMMER2017475,DEWAAL2022118348,koopman2021persuasive,vlek2016method} with only rare exceptions, like in~\cite{kyrimi2020incremental}, where the proposed explanation method was demonstrated to the clinicians.

The core idea of the proposed evaluation setup is to score the explanations first individually and then jointly. When scored individually, the participants are asked whether it is easy to follow the explanation and whether the explanation helped them understand how and why the probabilities in the target node were updated. Five answers from ``Strongly disagree'' to ``Strongly agree'' are proposed for selection. When scored jointly, the participants are first asked to choose the method that helped them better understand the reasoning process and then whether the combination of the proposed methods could help. For both joint questions, the participants have the option to answer ``None'' and ``The combination will not help much'' respectively. In both individual and joint scoring interfaces, the participant is provided with the option to leave textual feedback. See examples of all questions' interfaces (see Figures~\ref{fig:individual_questions},\ref{fig:joint_questions} in~\ref{sec:appendix_explanation}).

At the beginning of the explanation, the participant is presented with BN intuition, including a demonstrative example from Netica and bayesserver~\cite{bayesserver}. As a reference, we use the ASIA model (see Figure~\ref{fig:reasoning_sample}). The evaluation includes three scenarios, each corresponding to either predictive, evidential, or intercausal reasoning. To decrease the cognitive load, we randomly present only two scenarios to each participant, making sure that all scenarios get an equal number of demonstrations. Within each scenario, the \textit{baseline} explanation is always demonstrated first, and the \textit{fae} and \textit{incremental} explanations are demonstrated in a mixed order to prevent any order-specific bias.

We engage 25 anonymous participants from our professional network specializing in computer science and engineering research or development. Even though ASIA BN is related to the medical field, it was originally composed for demonstration purposes, so it may be properly understood without specialized knowledge in the medical area. None of them were aware of either which explanation method is ours or any other details of this study. 7 participants had not heard about BNs before the survey; 16 had only a general idea about BNs and only 2 identified themselves as experienced users of BNs. We take certain measures to make sure that the task is understood correctly and that the answers reflect the real properties of the perception of each particular participant. We analyze the textual comments from the participants and the time the person spent on each page. 
% First, we place an attention question at the end of the survey. This question is simply one explanation corresponding to one of three possible types that have clear inconsistencies (e.g. the target node stated in the task does not correspond to the target node used in the explanation). Figure~\ref{fig:foolish_fse} in Appendix~\ref{sec:appendix_questions} shows an example of such a question.
% 'no_exp': 7, 'heard': 16, 'experienced': 2

\subsection{Evaluation results}

\begin{figure*}[t!]
    \centering
    \includegraphics[width=\linewidth]{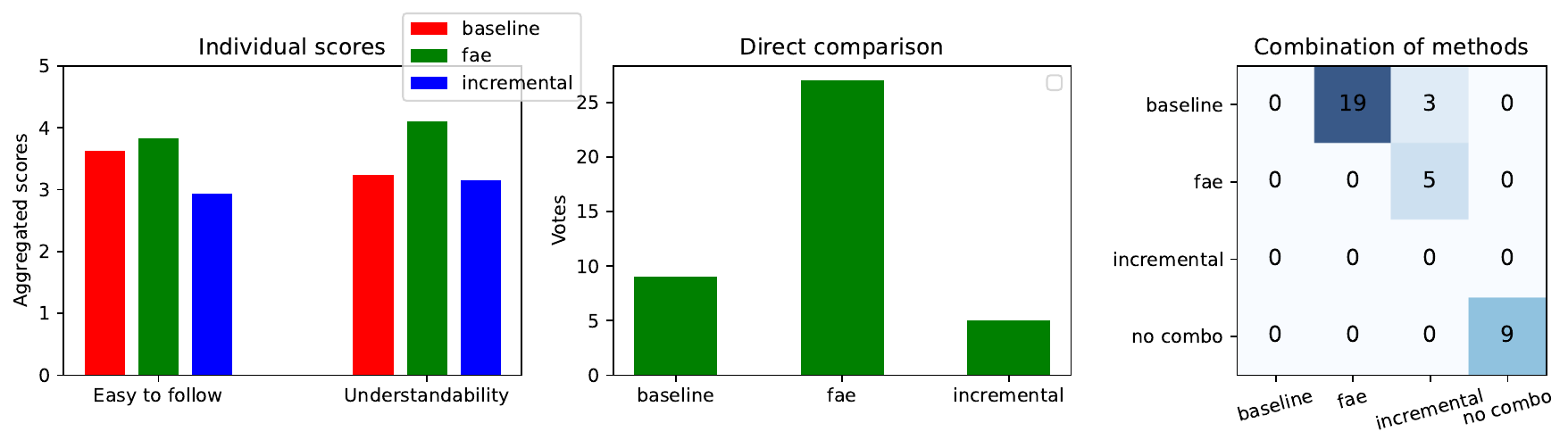}
    \caption{The results of the human-driven evaluation. The ``Direct comparison'' plot aggregates the answers about the explanation that helped to better understand the reasoning process. The ``Combination of methods'' plot shows the votes for the question about the potential use of the combination of the proposed methods (we show it as a diagonal matrix without duplication of the value for better readability of the plot).}
    \label{fig:eval_statistics}
\end{figure*}
% The ``Individual scores'' plot shows the aggregated answers to the questions about easiness to follow and understandability of the explanations. 

The proposed evaluation design turned out to be rather time-consuming and required a lot of cognitive power from participants. Its median passing time was 25 minutes. We manually examined both quality-control parameters described in the previous section and, finally, as an exclusion criteria, we dropped the answers from 4 participants whose textual comments indicated that they had not understood the task or who spent less than 30 seconds on any page of the task. 

The aggregated evaluation results are shown in Figure~\ref{fig:eval_statistics}. The ``Individual score'' plot shows the aggregated answers to the questions about each explanation type. The textual answers are mapped to numerical values from 1 (``Strongly disagree'') to 5 (``Strongly agree''). To verify the significance of the difference between the scores, we use a T-test with the null hypothesis of equality of all scores. The scores of easiness to follow the explanation flow are 3.63, 3.83, and 2.93 for \textit{baseline}, \textit{fae}, and \textit{incremental} correspondingly. The difference between \textit{baseline} and \textit{fae} turned out to be not statistically significant according to the T-test (p-value is 0.47, so we cannot reject the null hypothesis). This equal score is understandable because the \textit{baseline} method does not actually deliver any clarification and is pretty short. In terms of understandability, we can see that our method scored higher than the \textit{baseline} and \textit{incremental} methods (the scores are 3.24, 4.1, and 3.15). The results of the T-test let us reject the equality hypothesis between our method and \textit{baseline} and \textit{incremental} as far as p-values are 6E-4 and 1E-5, respectively, so the difference between these scores is statistically significant.

% We verified the significance of this difference with the T-test, and for both alternatives compared to our \textit{fae} method, the p-values were 6E-4 and 1E-5 for baseline and incremental, respectively, which means that the difference in this score is statistically significant. 

The ``Direct comparison'' plot shows the answers to the question about the explanation method, which helped to better understand the reasoning process. The final counts are 9, 27, and 5 for \textit{baseline}, \textit{fae}, and \textit{incremental} respectively. Our method clearly outperforms alternative ones. We verify the significance with a binomial test. The null hypothesis of this test is that the vote for \textit{fae} has equal chances of being selected by the participants compared to two other methods (namely that its probability is 1/3). The binomial test yields a p-value of 2E-5, which allows us to reject the null hypothesis.
% , s means that \textit{fae} is significantly more likely to be selected by the participants compared to the other two methods.

% more likely than 1/3 (namely more likely than for any one of three possible methods). This test yields a p-value of 2E-5 which means that \textit{fae} is significantly more likely to be selected by the participants compared to the other two methods.

% Second, we extract the votes from the participants that selected similar best algorithms for both scenarios they were presented with (the results are 1, 9, 1). From both of these points of view, our algorithm significantly outperforms the alternatives. 

Finally, the ``Combination of methods'' plot shows the answers to the question about the potential use of the combination of the proposed methods. We can see that the combination of \textit{fae} and \textit{baseline} received the most votes (19 of 36). This seems reasonable because the baseline answers the simple question ``What has changed, given the evidence?'' in a straightforward way and shows the dynamic of beliefs about all nodes of the BN, but does not provide any explanation. Whereas our approach explains the probability updates in a more detailed, step-by-step way. 

We also briefly analyzed the textual feedback from the participants. Some participants indicated that our algorithm delivers too many details of the reasoning, which may sometimes make perception more difficult and is not always necessary. However, at the same time, other participants found these details useful.  The incremental explanation algorithm turned out to be less understandable, partially because some participants failed to understand why some evidence was disregarded and also because the notion of factors that do not support the probability update is not clear. 

% The common feedback applicable to all types of explanations is hardly readable formatting of the text. Our initial belief was that using brackets around the nodes' names and states could make the text more readable and the significant information in the text easier to extract. However, for most of the participants, this format turned out to be difficult to read. The conclusion we can make from this feedback is that the potential explanatory tool should have explicit options to adjust the explanation to each particular user because it is hardly possible to select the format that would fit everyone's readability needs. 

Finally, most participants found the visual tips very helpful for understanding the explanation, which corresponded to our initial belief. Indeed, whereas the textual explanation may deliver important information about a BN's reasoning process, it could be particularly difficult to follow the explanation if there is no visual interpretation of it. A good idea may be to deliver such an explanation as an animated picture that could start from the initial state of the BN and then explicitly visualize the \textit{FA} flow (for our method) or highlight the described evidential and intermediate nodes (for incremental method).

The evaluation setup has certain limitations. First, the evaluation is performed on a basic BN of comparatively small size. Second, most participants in this evaluation possess significant knowledge in various fields of computer science or engineering, so their perception of the explanation may be biased. Further steps in the field of BN reasoning explanation have to be taken toward evaluation studies by expert users in the application domain (e.g., healthcare) with real-life BNs that are normally bigger than the one used in our study. However, such an evaluation study is worth a standalone paper~\cite{BUTZ2022102438}, so we leave this task for future work.

% Another issue that is missing in the paper is the evaluation by expert users in the application domain (e.g. healthcare) and not by technical users, whose view of what is useful in terms of explainability may be biased by their technical knowledge (even if not specifically about BN).

\section{Conclusions}

In this work, we introduce a novel approach to the textual explanation of BN inference, which is based on the notion of a factor argument—abstract representations of the flow of information that tangle observations with variables of interest. We present how to use this formalism for the content determination stage of natural-language explanations of approximate reasoning in BNs and, in particular, to decide when to present considerations jointly or separately. We experimentally show that our proposed approach accurately approximates the message-passing algorithm. Finally, we perform a human-driven evaluation using the medical domain BN of the proposed natural-language explanations by comparing it with another explanation method and with the baseline description of the BN. The results of the evaluation show that our method is significantly more understandable than the compared method.

\section{Data statement}

The BNs used for the experiments in this work are available on bnlearn website~\cite{bnlearn}.

\begin{acknowledgments}
  This research was funded by the European Union's Horizon 2020 research and innovation program under the Marie Skodowska-Curie grant agreement No 860621, and the Galician Ministry of Culture, Education, Professional Training, and University and the European Regional Development Fund (ERDF/FEDER program) under grants ED431C2018/29 and ED431G2019/04.  
\end{acknowledgments}

%% Define the bibliography file to be used
\bibliography{literature}
% \bibliography{sample-ceur}

%%
%% If your work has an appendix, this is the place to put it.
\appendix

\FloatBarrier

\section{Human-driven evaluation survey}
\label{sec:appendix_explanation}

\begin{figure}[t]
    \centering
    \includegraphics[scale=0.6]{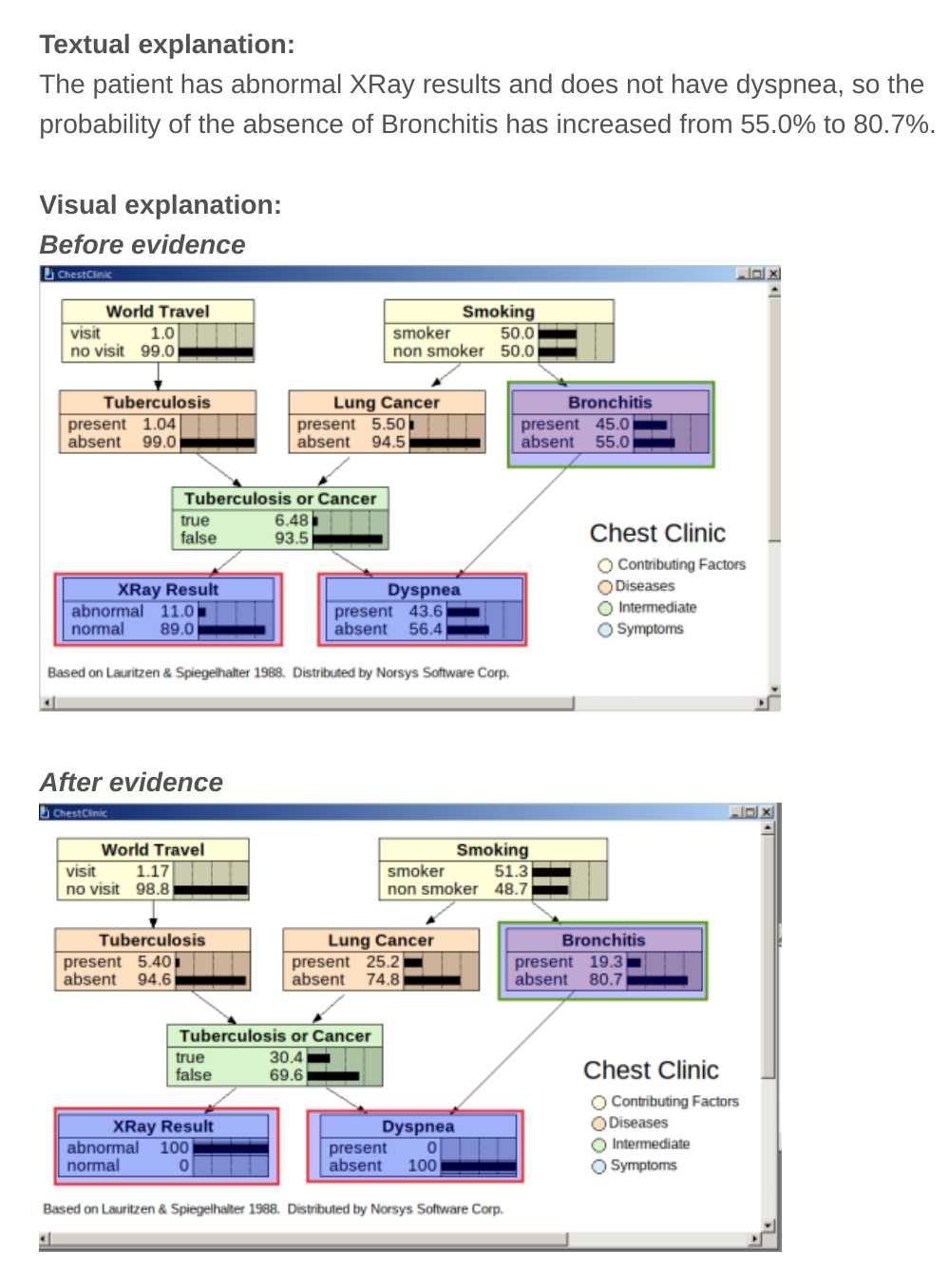}
    \caption{Example of baseline explanation in the interface of the human-driven evaluation survey}
    \label{fig:baseline_example}
    \end{figure}

\begin{figure}[]
    \centering
    \includegraphics[scale=0.8]{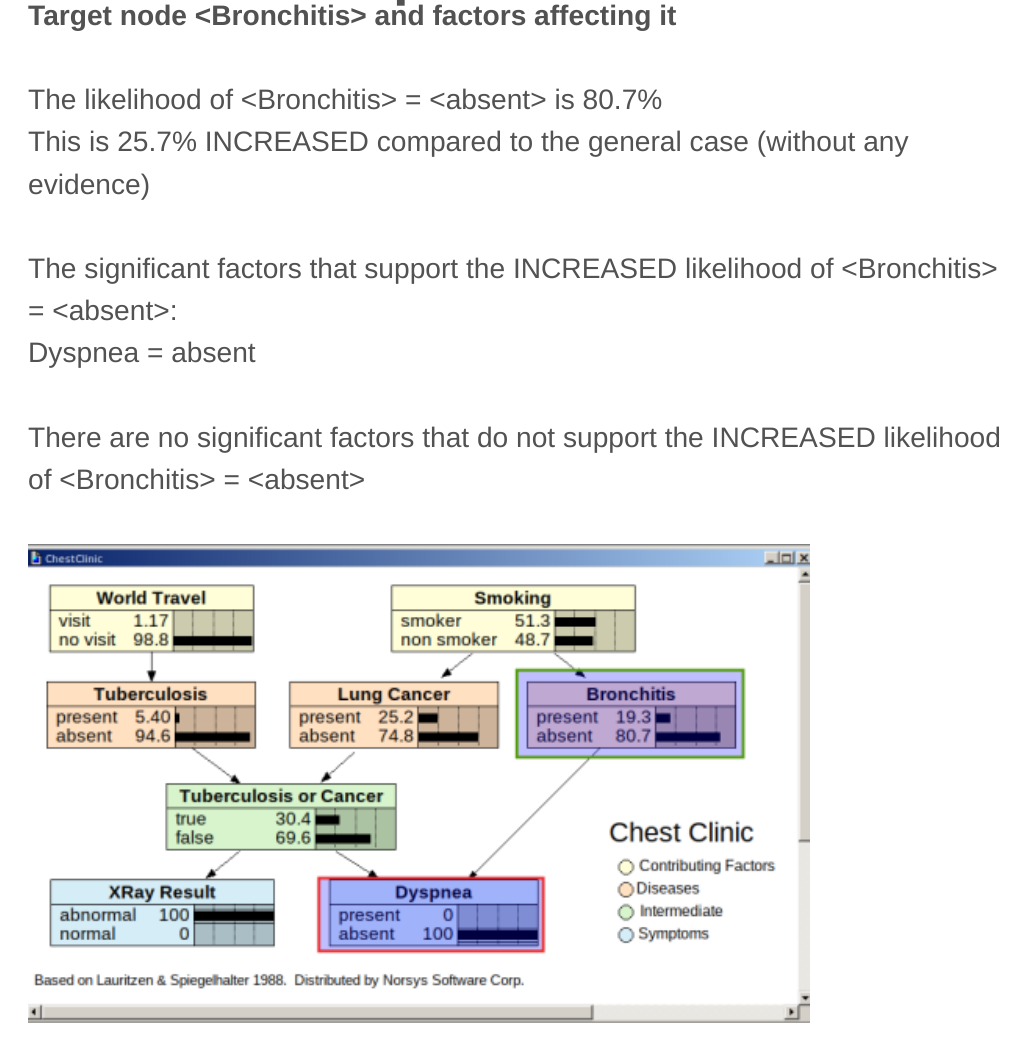}
    \caption{Example of incremental explanation (first explanation stage described in~\cite{kyrimi2020incremental}) in the interface of the human-driven evaluation survey}
    \label{fig:increm_example}
    \end{figure}

\begin{figure}[]
    \centering
    \includegraphics[scale=0.8]{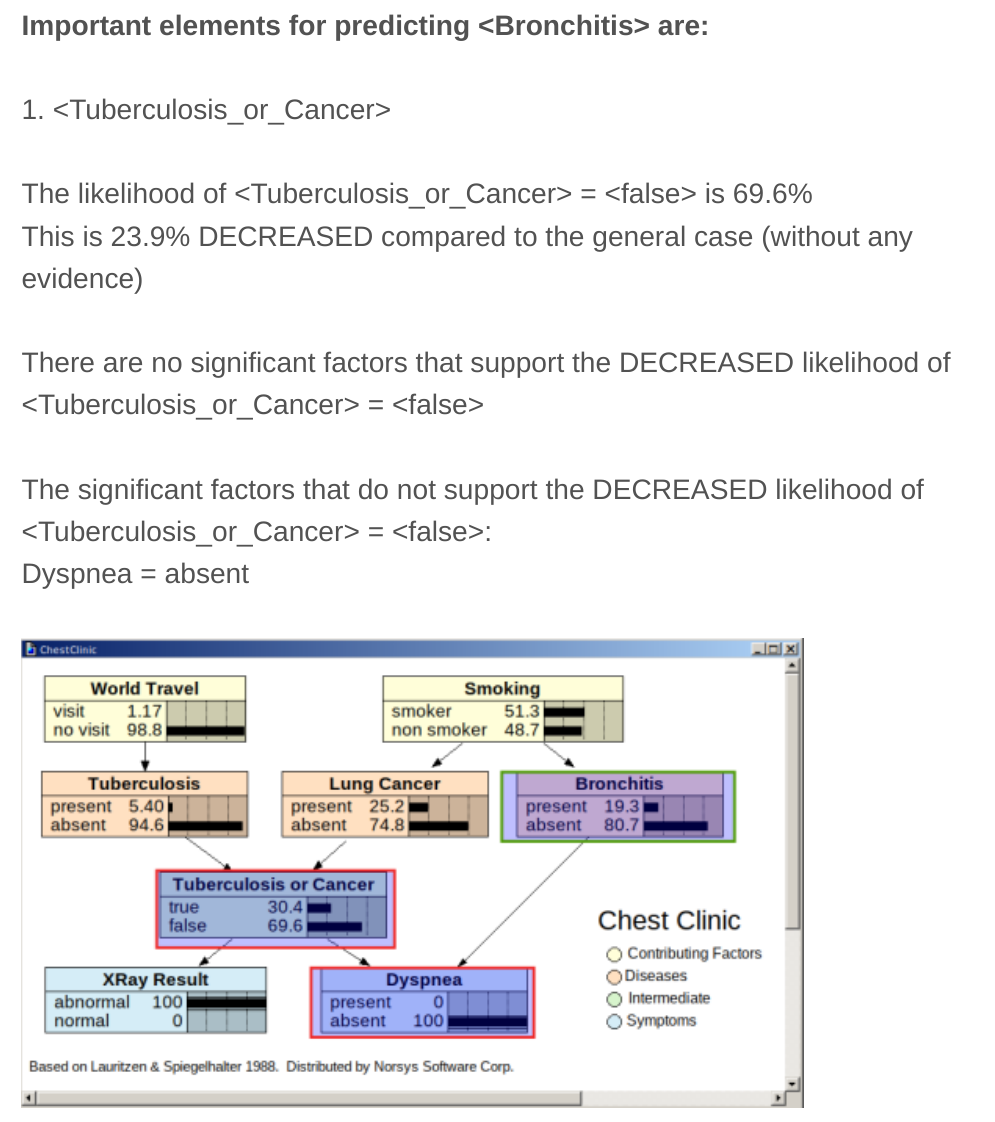}
    \caption{Example of incremental explanation (second and third explanation stages described in~\cite{kyrimi2020incremental}) in the interface of the human-driven evaluation survey}
    \label{fig:increm_example_23}
    \end{figure}

\begin{figure}[]
    \centering
    \includegraphics[scale=0.8]{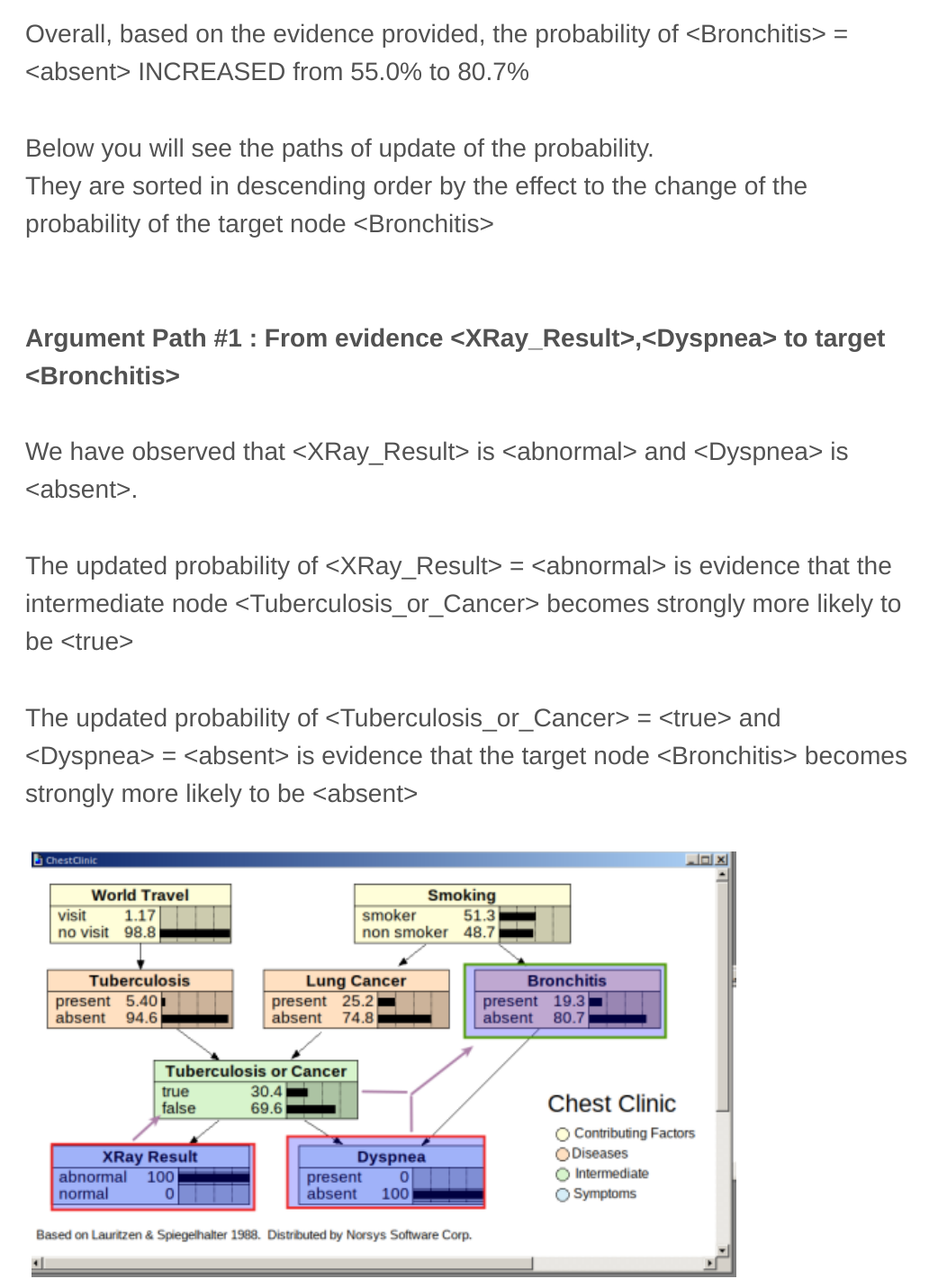}
    \caption{Example of our explanation method in the interface of the human-driven evaluation survey}
    \label{fig:fse_example}
    \end{figure}

% \begin{figure}[]
%     \centering
%     \includegraphics[scale=0.8]{foolish_fse.pdf}
%     \caption{Example of attention question in the interface of the human-driven evaluation survey}
%     \label{fig:foolish_fse}
%     \end{figure}

% \FloatBarrier

% \section{Examples of questions in human-driven evaluation survey}
% \label{sec:appendix_questions}

\FloatBarrier

\begin{figure}[t]
    \centering
    \includegraphics[scale=0.55]{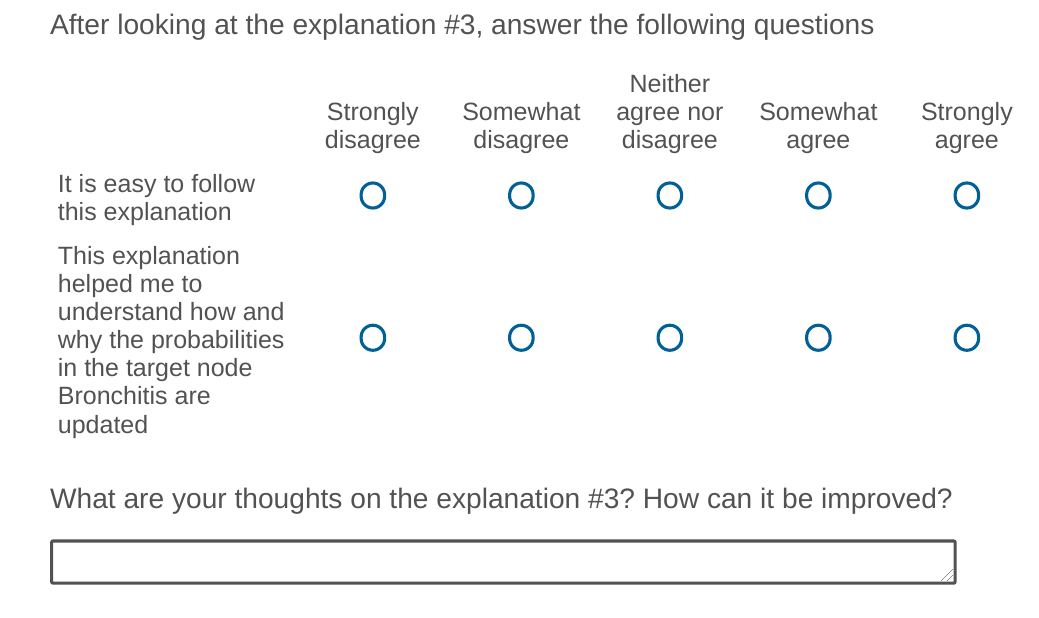}
    \caption{Example questions about definite explanation method in the interface of the human-driven evaluation survey}
    \label{fig:individual_questions}
    \end{figure}

\begin{figure}[t]
    \centering
    \includegraphics[scale=0.6]{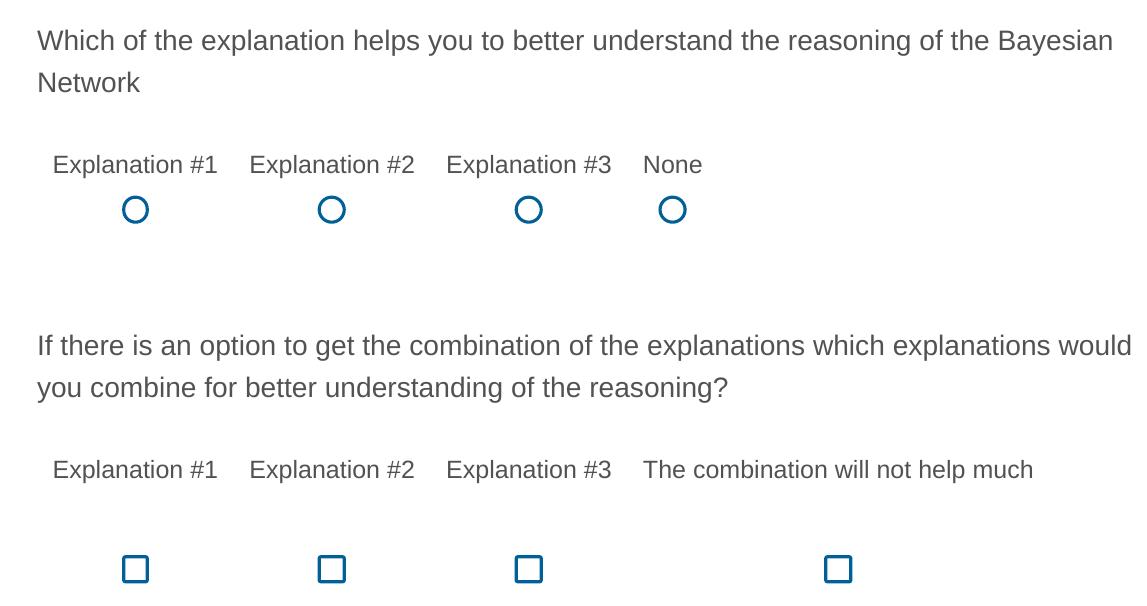}
    \caption{Example questions about all explanation methods within one case in the interface of the human-driven evaluation survey}
    \label{fig:joint_questions}
    \end{figure}

\end{document}